\documentclass[acmsmall, screen]{acmart}
\acmJournal{THRI}

\usepackage{colortbl}
\usepackage{multirow}

\AtBeginDocument{%
  }

\setcopyright{acmlicensed}
\copyrightyear{2018}
\acmYear{2018}
\acmDOI{XXXXXXX.XXXXXXX}
\acmBooktitle{Woodstock '18: ACM Symposium on Neural Gaze Detection,
 June 03--05, 2018, Woodstock, NY}
\acmISBN{978-1-4503-XXXX-X/2018/06}

\begin{document}

\title[Grounding and Communication Effort in Human-Robot Collaboration]{Grounding and Communication Effort in Human-Robot Collaboration: The Influence of Robot Response Strategies}

\author{Yan Zhang}
\affiliation{%
  \department{School of Computing and Information Systems}
  \institution{University of Melbourne}
  \city{Melbourne}
  \state{VIC}
  \country{Australia}}
\email{yan.zhang.1@unimelb.edu.au}
\orcid{0000-0003-2142-5094}

\author{Jarrod Knibbe}
\affiliation{%
  \department{School of Electrical Engineering and Computer Science}
  \institution{The University of Queensland}
  \city{Brisbane}
  \state{QLD}
  \country{Australia}}
\email{j.knibbe@uq.edu.au}
\orcid{0000-0002-8844-8576}

\author{Jorge Goncalves}
\affiliation{%
  \department{School of Computing and Information Systems}
  \institution{University of Melbourne}
  \city{Melbourne}
  \state{VIC}
  \country{Australia}}
\email{jorge.goncalves@unimelb.edu.au}
\orcid{0000-0002-0117-0322}

\author{Wafa Johal}
\affiliation{%
  \department{School of Computing and \\Information Systems}
  \institution{University of Melbourne}
  \city{Melbourne}
  \state{VIC}
  \country{Australia}}
\email{wafa.johal@unimelb.edu.au}
\orcid{0000-0001-9118-0454}

\renewcommand{\shortauthors}{Zhang et al.}

\begin{abstract}
Effective human-robot collaboration depends critically on communication. Current robot design often places the communicative work, e.g. grounding (establishing mutual understanding and repairing misunderstandings), on the robot, reflecting a seemingly human-centred goal of minimising human effort. Research on human-human teams, however, suggests that communication is a shared accomplishment, raising the question of whether a robot carrying more of the communicative work actually makes communication feel less effortful for humans. We investigated this question in a mixed-design empirical study with 32 participants, examining how explicitness and correctness of a robot's responses affect participants' perceived grounding and communication effort. The robot responded to participants' instructions either explicitly or implicitly (with or without verbally confirming its understanding), and either correctly or incorrectly interpreted the participants' intent. All participants completed tasks of two severity levels, in which errors carried either low or high consequences. The results show that perceived effort was associated with correctness. Participants experienced correct implicit responses as the least effortful, whereas incorrect implicit responses shifted the burden of detecting and repairing the misunderstanding onto them. In high-severity tasks, however, explicit responses changed this pattern. Participants valued the communication effort shared by both partners over minimising it. These findings suggest a principle of balanced communication effort, context-dependent and shared rather than always minimised, with implications for the design of collaborative robots.
\end{abstract}

\ccsdesc[500]{Human-centered computing~Empirical studies in HCI}
\keywords{Human-Robot Collaboration, Communication, Grounding, LLM, Empirical Study}



\maketitle

\section{Introduction}
Consider a person working at a desk in the early evening who remarks to a robot assistant, ``It's dark in here.'' The remark is an indirect request whose meaning depends on context, that is, it means turn on the lamp if the person is settling in to keep working at night. If the same remark is made in the early morning, however, it means open the curtains to let the daylight in. Suppose the robot picks the plausible yet contextually wrong reading and acts on it immediately, without confirming its understanding. By design this is efficient, since it spares the human any extra exchange. In effect, it shifts the real work onto the human, who must notice the mistake, undo it, and re-explain the request in a more direct but less natural way. The robot was built to reduce human effort, but it has produced more. This scenario exposes a tension at the core of human-robot collaboration (HRC). Robots are increasingly designed to work alongside humans with the goal of reducing human effort~\cite{lorenzini_ergonomic_2023, jung_what_2019, jonnavittula_sari_2024}, and this goal has driven collaborative systems that prioritise task efficiency and human effort minimisation, favouring the kind of behaviour the robot above displays. But human collaboration is not merely a matter of minimising individual effort. It is an interactive process mediated mainly through verbal communication, in which effort is distributed between partners~\cite{clark_coordinating_2005}. For robots intended to operate as flexible, communicative partners in open-ended settings, rather than as fixed-function machines performing repetitive tasks within predefined parameters, an efficiency-only framing obscures this interactive nature. It assumes that effort can be straightforwardly offloaded to robots, when in practice minimising humans' effort and maintaining mutual understanding can pull in opposite directions, as the opening scenario shows. Understanding how verbal communication actually shapes the dynamics of effort and coordination in HRC therefore requires looking beyond efficiency-oriented assumptions.

Understanding how communication shapes collaboration involves examining how grounding is established and how effort is distributed. Grounding is the process by which collaborators establish common ground, defined as a set of shared contextual knowledge, beliefs, and assumptions~\cite{clark_grounding_1991}. In human-human collaboration (HHC), grounding is central to success because it enables interlocutors to coordinate actions and work towards mutual understanding and shared goals through the exchange of intent~\cite{clark_using_1996}. Research has shown that humans follow the principle of least collaborative effort, seeking to minimise the overall effort of the team rather than the effort of any individual member~\cite{clark_referring_1986}. Current robot design takes a different path. It is driven by the goal of minimising the human's individual effort, assuming that offloading work to the robot always improves collaboration. For robots that communicate with people and act with a degree of autonomy in shared physical spaces, whether and how the team-level principle from HHC applies remains unclear, especially regarding how robots' communicative behaviours influence grounding and the distribution of communication effort.

Central to effective grounding is how robots interpret human intent. People frequently convey intent implicitly through indirect requests, which go beyond literal meaning and require interpretation through context~\cite{clark_referring_1986}. 
The opening scenario illustrates this, as the same remark calls for turning on the lamp in the evening but for opening the curtains in the morning, and only context can disambiguate the two.
The naturalness and flexibility of indirect requests make them well-suited for HRC~\cite{liu_review_2019, obaigbena_ai_2024_2}, and empirical studies suggest that humans use them with robots at similar frequencies as with other humans~\cite{williams_thank_2018}. Robots capable of interpreting such contextualised intent have been shown to enhance grounding~\cite{bennett_differences_2017, lee_receptionist_2010, zhang_can_2025, zhang_implicit_2025}. However, the inherent ambiguity of indirect requests introduces misunderstandings unavoidable in real-world HRC. Misinterpretations not only reduce team fluency and goal alignment but may fundamentally shape how grounding is achieved and how communication effort is distributed between human and robot~\cite{zhang_can_2025}. Therefore, the impact of a robot's correctness in interpreting contextualised intent on grounding and effort distribution requires further investigation.

Whether a misinterpretation becomes a costly failure or a quickly repaired hiccup depends on how the robot communicates its understanding. Researchers have explored approaches where robots respond explicitly, confirming what they understood and what they intend to do~\cite{whitney_reducing_2017, dai_think_2024}. Such explicit responses may allow errors to surface earlier, and strengthen trust by providing clarity about the robot's belief state~\cite{smith_exploring_2024, alhaji_trust_2021}. This clarity comes at a price, as explicit responses demand attentional and communicative effort from the human. When a robot responds implicitly and simply acts~\cite{cui_no_2023, thierauf_this_2024}, it may minimise immediate effort, but a misinterpretation can proceed until its consequences appear, leaving the human to detect and repair the error. Explicitness is thus more than a stylistic choice, as it shapes whether misinterpretations are caught or missed. This interplay between response explicitness and interpretation correctness has received limited attention in HRC.
Does the value of explicit confirmation depend on whether the robot misinterprets? Do implicit responses consistently achieve effort economy across different collaborative contexts?

The answers to these questions may further depend on what is at stake when errors occur. Since a task itself does not carry severity, we consider the severity of error consequences, that is, how harmful the potential negative outcomes of a mistake are. Severity of consequences has been identified as a moderator of human perceptions in HRC~\cite{hancock_meta-analysis_2011, bhat_value_2024}, and prior work suggests that language use in HRC is task-dependent~\cite{zhang_can_2025}. There is also a mechanistic reason to expect moderation. When the consequences of an error are severe, the cost of a missed misunderstanding may outweigh the cost of redundant confirmation, shifting the effort calculus away from pure minimisation. Moreover, the perceived severity of negative outcomes has been found to be associated with how people attribute blame after errors in human-AI interaction, both to the artificial system and to themselves~\cite{leichtmann_moral_2024}.
Therefore, task severity is a critical moderating factor warranting investigation when examining robots' communicative responses in collaborative tasks.

Building on this literature, we investigate how the \emph{correctness} (correct vs. incorrect) of a robot's contextual intent interpretation and the \emph{explicitness} (implicit vs. explicit) of its responses influence grounding and effort in human-robot physical collaboration, and how these effects are moderated by \emph{task severity} (high vs. low). Specifically, we explore the following questions:
\begin{description}
    \item [RQ1.1] How does robots' contextual intent interpretation influence human-robot collaboration?
    \item [RQ1.2] To what extent does robots' communicative explicitness moderate the effect of contextual intent interpretation on human-robot collaboration?
    \item [RQ2.1] How does robots' communicative explicitness influence human-robot collaboration?
    \item [RQ2.2] To what extent does robots' contextual intent interpretation moderate the effect of communicative explicitness on human-robot collaboration?
    \item [RQ3] To what extent does task severity moderate the effects of communicative explicitness and contextual intent interpretation on human-robot collaboration?
\end{description}

To address the above questions, we conducted a lab study in which 32 participants collaborated with two robots on two physical tasks that differed in the severity of error consequences. We measured participants' experience of grounding, their perceived communication effort, and the perceived robot communication effort. We complemented these quantitative measures with a qualitative analysis of participants' accounts, which forms a central part of our findings.

This paper makes three contributions. First, it provides empirical evidence of how correctness and response explicitness shape grounding and the distribution of communication effort in physical HRC. Second, through a qualitative analysis of interviews about participants' collaboration with the robots, complemented by quantitative measures, it reveals how the severity of tasks moderates the experience of communicative effort, clarifying when participants described additional effort from the robot or themselves as reassuring and worthwhile, and when they experienced it as tiring and unnecessary. Third, drawing on these findings, we propose the principle of balanced communication effort in HRC, which extends the least collaborative effort principle from HHC by showing that effort should be adjusted to context and stakes, not unconditionally minimised or offloaded to robots. In summary, these contributions invite a reconsideration of the assumption that collaboration improves simply by shifting effort onto robots, and they point towards a design paradigm in which collaborative robots act not only as efficiency-optimised task executors but as adaptive communicative teammates capable of navigating the interactive complexities of joint work.

\section{Related Work}
\subsection{Grounding, Uptake, and Effort in Communication}
Grounding has long been discussed in human-human communication and has been acknowledged as closely tied to collaborative effort. Grounding is commonly defined as the process by which interlocutors establish and update common ground in conversation~\cite{clark_grounding_1991}. This process consists of two phases: the presentation phase, in which the speaker offers an utterance for the addressee to consider, and the acceptance phase, in which the addressee signals that they believe they understand. 
In this process, the speaker and addressee implicitly work together to establish and maintain the mutual belief that the addressee has understood the speaker's intent. If this belief is in doubt, repair or clarification is triggered. This aligns with the notion of uptake, whereby the speaker ensures that the addressee has actively understood their intent, highlighting the inherently interactive and co-constructed nature of successful communication~\cite{austin_how_1975}. 

To achieve uptake, grounding requires actions. Researchers have identified seven grounding acts: initiate, continue, request acknowledgement, acknowledgement, repair, request repair, and cancel~\cite{traum_computational_1995, visser_model_2014}. Initiate, request acknowledgement, and cancel acts are produced by the speaker, who respectively introduces an utterance, solicits a response, or withdraws a previously established belief. Acknowledgement is produced by the addressee to accept the speaker's utterance. Repair and request repair can be initiated by either interlocutor, soliciting collaborative contributions to re-establish common ground. Communication is not limited to verbal dialogues, physical actions can also contribute to grounding~\cite{zheng_putting_2024}. Researchers have shown that simply comprehending sentences implying actions engages motor systems, with the action-sentence compatibility effect demonstrating that understanding language is intrinsically tied to embodied action~\cite{glenberg_grounding_2002}. This suggests that grounding is closely linked to physical collaboration, making it a worthwhile focus of investigation with embodied agents. On the other hand, grounding-relevant signals need not be intentionally designed as such. Because humans continuously interpret a robot's behaviour, even non-communicative actions can enter the grounding process. Evidence showed that users' understanding of a robot was continually reshaped by their interpretations of its movements and signals, regardless of whether these interpretations matched how the robot actually worked~\cite{leichtmann_teaming_2025}. In summary, these perspectives highlight that grounding requires effort from both sides, and that any observable behaviour of an embodied agent may feed into common ground, whether designed to or not.

Accordingly, communication effort has consistently been discussed alongside grounding~\cite{clark_referring_1986}. In conversation, speakers and addressees attempt to minimise collaborative effort. This involves a trade-off between offering barebones descriptions~\cite{clark_using_1996}, which invite the addressee to contribute interpretive work and provide quick acknowledgement, and refashioning a description if it is rejected~\cite{clark_grounding_1991, clark_referring_1986}. This illustrates the collaborative nature of dialogue and the principle of least collaborative effort. Evidence suggests that humans follow this principle: for example, if one speaker uses a particular syntactic structure or lexical choice and is understood, the other is more likely to unconsciously reuse that form, thereby aligning their language as a way of minimising effort~\cite{pickering_toward_2004}. Communication effort has been regarded as evidence of participants' contributions to the processes of collaboration and grounding~\cite{clark_contributing_1989}. Despite substantial insights into mechanisms of grounding and communication effort from HHC, the implications for and empirical studies on collaborative robots remain limited.

\subsection{Effort Allocation between Human and Robot}
Previous work has investigated communication effort together with grounding. An empirical study found that humans developed false beliefs about common ground when robots provided low communication effort, whereas they reached real grounding with robots with high effort, which revealed robots' internal representation~\cite{chai_collaborative_2014}. Most studies have focused on making robots take the majority of the effort in human-robot communication. ~\citet{devin_implemented_2016} proposed a system for robots' theory of mind, the ability to reason about others' perceptions, beliefs, and goals, enabling robots to evaluate, adapt to, and confirm humans' intent, thereby reducing humans' effort. Moreover, robots need to detect referential ambiguity and recover from misunderstanding to improve dialogue efficiency~\cite{marge_miscommunication_2019}. Robots are also expected to be more proactive in contributing to mutual understanding rather than a passive executor~\cite{belhassein_addressing_2022}. Some researchers equipped robots with large language models (LLMs) that generate both verbal and gestural communication, taking responsibility for confirming humans' ambiguous instructions~\cite{allgeuer_when_2024}. 
However, there can also be too much communication. Explicitly conveying details that are already obvious to expert users or users familiar with the system can distract from the task and annoy them. Robot instructions elaborated for novices proved redundant for experts and reduced the efficiency of information exchange~\cite{torrey_effects_2006}, and confirmations of already-correct interpretations similarly cause frustration, with users expecting robots to drop them as familiarity grows~\cite{reimann_survey_2024, smith_exploring_2024}. Thus, although placing more burden on robots intuitively should reduce humans' effort, the insights from HHC and these findings suggest important caveats. Whether robots' extra communication effort robustly reduces humans' effort requires further empirical evidence.

On the other side, humans' communication efforts are commonly used for requesting and training robots. Some studies found that humans tended to adjust their word choices to align with robots' expression, indicating humans taking on grounding effort by tailoring their speech to robots' communication ability~\cite{iio_lexical_2009, iio_lexical_2015, brandstetter_persistent_2017}. Empirical evidence showed humans adjusted their pragmatic usage by shifting from indirect speech acts to direct ones to ensure robots' understanding, which demanded more mental workload to keep the interaction fluent~\cite{zhang_can_2025}. Moreover, researchers found that humans naturally increase communicative effort when facing mismatched perception with robots~\cite{chai_collaborative_2016}.

A few studies looked at shared efforts. MICoBot implemented mixed-initiative dialog under real-world scenarios, which enabled bi-directional intent negotiating for effective collaboration with various users~\cite{yu_mixed-initiative_2025}. Moreover, researchers found humans tended to use shorter explanations toward robots that they perceived as capable, reflecting the same strategy of minimising overall effort when collaborating with humans~\cite{kiesler_fostering_2005}. Similarly, ~\citet{kontogiorgos_towards_2020} proposed that robots should also follow the principle of least collaborative effort when repairing mismatched common grounds with humans. However, the above literature lacks consideration of situational stakes and empirical evidence on humans' and robots' grounding and communication effort in physical collaborative settings.  

\subsection{Explicitness and Intent Interpretation in Human-Robot Collaboration}
Previous studies revealed that to achieve grounding in HRC, verbal communication is a critical channel. Researchers used a computational model for referring expression generation to help ground language~\cite {chai_collaborative_2016}. The robots made an extra effort to explicitly express their understanding via multi-step iteration, which improved true grounding performance when the perceptual mismatch was high. 
This study focused on grounding the references of words. The human and the robot perceive the same scene differently, so they must establish a shared mapping between the human's labels and the objects in the robot's perception. For example, when the human refers to an object as ``Eric,'' the robot must verify that this name corresponds to the object it perceives as an orange bottle, and only once both partners agree that ``Eric'' denotes that particular bottle is the reference grounded. Researchers also argued that confirmation is not always beneficial. When the robot's interpretation of the human's intent is already correct, requesting confirmation is unnecessary and can lead to additional interaction cost, for example in interactive reinforcement learning settings~\cite{reimann_survey_2024}.
With the development of LLMs, robots have become more effective in reaching mutual understanding with humans' linguistic variability when referring to an unknown object in the real world~\cite{jiang_llms_2024, wanna_lets_2025, gao_physically_2024}. LLMs greatly increase robots' ability to understand which objects humans are referring to, thereby reducing the need and frequency of explicit confirmations. However, the ability to acquire linguistic variability is not sufficient to fully understand human intent in highly collaborative tasks. Humans' action-oriented intents require interpretation in relation to the context~\cite{zhang_can_2025}. Therefore, confirmation of humans' intent remains an indispensable aspect of HRC.

There are different ways to clarify and confirm the correctness of a robot's understanding with humans. Researchers have explored how robots verbally confirm with humans~\cite{smith_exploring_2024, grasse_speech_2021}, as well as the use of verbal and deictic gestures to ensure accurate interpretation amid the noise of natural language and human gestures~\cite{whitney_reducing_2017}. Explicit confirmation can increase trust and task success, but may also cause a degree of annoyance. In long-term collaboration, researchers found that robots' confirmation strategies may shift from explicit confirmation, to acting directly with accompanying verbal interaction, and ultimately to completing actions without asking~\cite{smith_exploring_2024}. Moreover, robots are expected to explicitly initiate confirmation with humans when intent recognition fails~\cite{lyu_signaling_2025}. While these approaches focus on task success, less attention has been paid to the collaborative process itself, particularly the distribution of effort and humans' subjective perception of the robot's contribution as a request addressee.

Previous work has also investigated how a robot uses actions to implicitly communicate with humans. Researchers stated robot actions included both functional and communicative dimensions~\cite{knepper_implicit_2017}. When robots take an unexpected action in the context of the task, humans may interpret it as a signal of misunderstanding, and vice versa~\cite{knepper_implicit_2017, liang_implicit_2019}. Other studies have employed robots' movement speed to convey uncertainty and internal confidence~\cite{hough_its_2017, schombs_robot-assisted_2024}. These findings suggest that signals for grounding may possibly be communicated implicitly through physical action, rather than relying on explicit confirmation only. However, such implicit communication may also risk misinterpretation. Because humans continuously interpret robot behaviour, including signals that were not intended communicatively, they may draw incorrect inferences about the robot's understanding or state. For instance, findings showed that users initially read a robot's status lights as feedback on whether their input gestures had been detected, when the lights merely indicated whether the robot was moving or standing~\cite{leichtmann_teaming_2025}. This creates a trade-off between efficiency and the potential frustration of misinterpretation, versus high accuracy accompanied by the potential annoyance of constant explicit confirmation.

In summary, prior work has linked grounding to communication effort, yet it remains unclear whether shifting this burden to robots is the most suitable approach in HRC, especially under varying situational stakes. While explicitness in communicating the interpreted intent can improve task success, it also risks negative human perceptions. Empirical evidence is therefore needed on how robots' explicitness and intent interpretation correctness shape grounding and communication effort across different levels of task severity in HRC.

\section{User Study}
To investigate the influence of robots' explicitness, correctness, and the moderation from task severity, we designed and implemented a multi-agentic system to support robots' verbal communication and object manipulation under different conditions. We conducted an in-lab user study in a triad setup consisting of one participant and two robots.

The experiment followed a mixed factorial design, with \emph{Explicitness} (Implicit vs. Explicit) and \emph{Correctness} (Correct vs. Incorrect) as between-subject main factors, and \emph{Severity} (High vs. Low) serving as a within-subject moderator factor. Crossing the two between-subject factors produced four experimental groups, namely $Implicit\times Correct$ (IC), $Implicit\times Incorrect$ (II), $Explicit\times Correct$ (EC), and $Explicit\times Incorrect$ (EI). Each participant was randomly assigned to one of these four groups and completed both the high-severity and the low-severity task in a counterbalanced order, yielding a 2 (Explicitness, between) $\times$ 2 (Correctness, between) $\times$ 2 (Severity, within) mixed design. \autoref{tab:conditions} summarises how the Collaborator's behaviour was varied in each condition.
The study employed a mixed-method design~\cite{creswell_mixed-method_1999}, collecting quantitative data through questionnaires on the perception of grounding and communication effort, as well as qualitative data from semi-structured interview responses. Further details of the experimental design are provided in the following sections.

\begin{table}[t]
\centering
\caption{Overview of the four between-subject conditions. Each participant experienced one condition across both tasks. In all conditions, Severity (High vs. Low task) varied within subjects.}
\label{tab:conditions}
\begin{tabular}{llll}
\hline
Condition & Confirms before acting & Wording of object references & Intent interpretation \\
\hline
IC & No & Mirrors participant's phrasing & Correct \\
II & No & Mirrors participant's phrasing & Plausibly incorrect\\
EC & Yes & Precise descriptions & Correct\\
EI & Yes & Precise descriptions & Plausibly incorrect \\
\hline
\end{tabular}
\end{table}

\subsection{Apparatus and Setup}
\begin{figure*}
    \centering
    \includegraphics[width=0.8\textwidth]{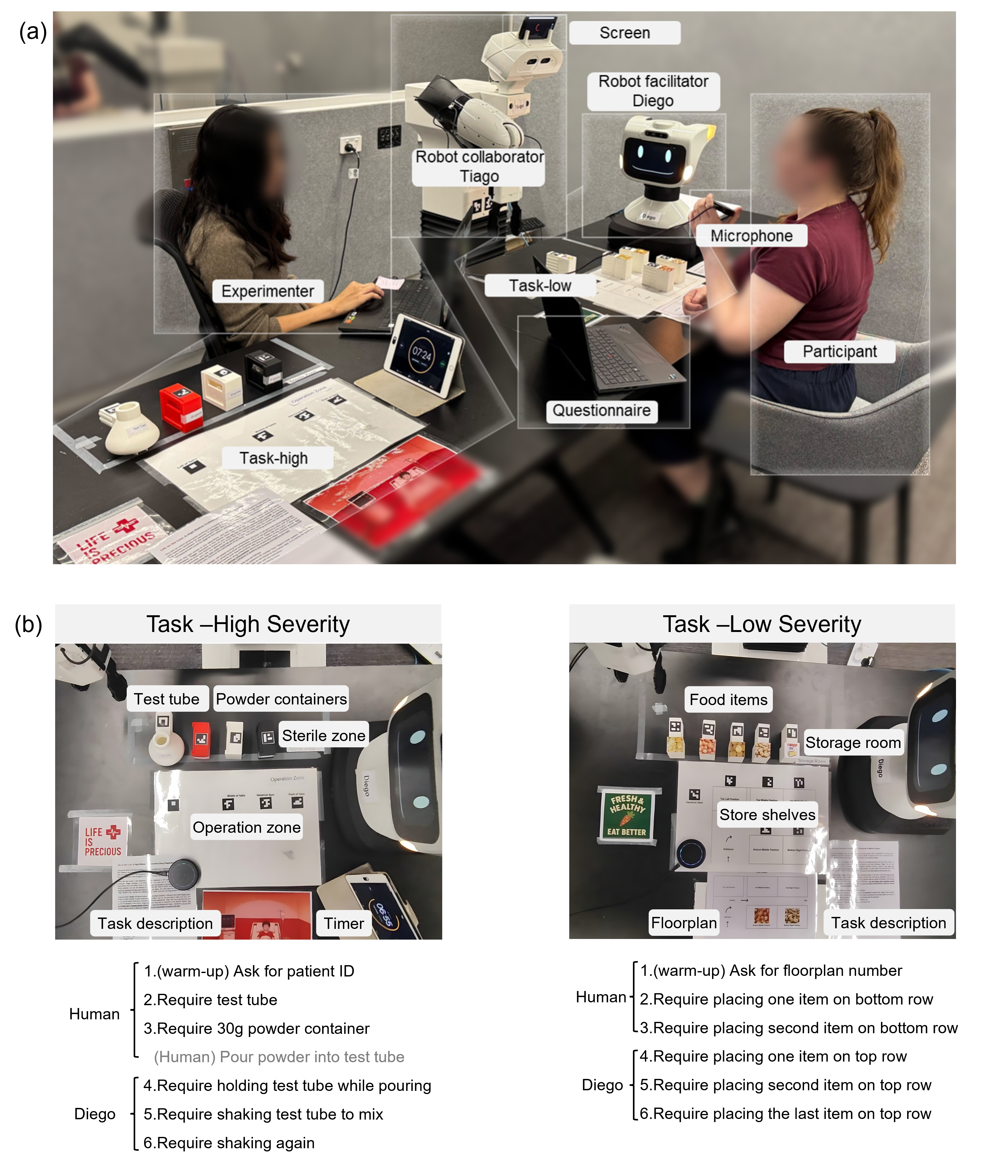}
    \caption{(a) This figure illustrates the experimental setup. (b) The high-severity task is shown on the left and the low-severity task on the right, with the distribution of task steps displayed below. Participants had no access to the Facilitator's steps during the experiment.}
    \Description{The figure has two parts. In part a, the experiment setup is shown. An experimenter sits on one side of the table with a robot collaborator called Tiago, which has an arm and is used for the collaboration tasks. A small screen is mounted on Tiago's head. On the other side sits a participant. Another robot called Diego, which has a screen face and acts as the robot facilitator, is on the desk between the participant and Tiago. A microphone is placed near the participant, and a laptop for questionnaires is on the table. They are currently sitting at the table for a high-severity task. On the side is the table for the low-severity task. In part b, two close-up overhead views show the tasks. On the left is the high-severity task, which is a medical task. Items include a test tube, three powder containers with red, white and black colours, a sterile zone, and an operation zone. A timer and a task description sheet are also present. The task steps are listed below: the human asking for the patient ID, then requesting a test tube, requesting a powder container, and pouring the powder into the tube. Then Diego asks Tiago to hold the test tube steady while pouring, and shake the tube twice to mix. On the right is the low-severity task, which is a grocery store stocking task. Items include food boxes, a storage room area, and labelled store shelves, along with a floorplan and task description sheet. The task steps are listed below: the human asking for a floorplan number, then requiring Tiago to place items on the bottom row, and then Diego asking Tiago to place items on the top row of the shelves.}
    \label{fig:exp}
\end{figure*}

The experiment was conducted in a quiet, isolated room, with two robots. Participants engaged in physical collaborative tasks with a \textbf{Robot Collaborator} serving as their partner. This Robot Collaborator is Tiago, a mobile manipulator robot with anthropomorphic features, including a head, neck, torso, arm, and a moving base, making it well-suited for human-robot interaction (HRI) research and capable of object manipulation~\cite{pages_tiago_2016}. A phone was mounted on the Robot Collaborator's head using a custom 3D-printed holder, serving as an external display (see~\autoref{fig:exp}). Two animations were shown on this screen to indicate whether the Robot Collaborator was listening to the participant's speech or processing input, thereby preventing the robot from listening to itself or missing participants' speech. The Robot Collaborator and the participant were positioned on opposite sides of a table, which served as the shared workspace.

A second robot, Diego, was included in the experiment as a \textbf{Robot Facilitator} to issue three indirect requests to the Robot Collaborator following the participant's task steps. This design choice was necessary because participants' verbal utterances were not restricted, in order to preserve natural interaction. Consequently, it could not be guaranteed that a given participant would spontaneously produce an indirect request. A second and scripted robot therefore ensured that the Collaborator's responses to indirect requests were reliably triggered independent of the participant's own speech. This design ensured that all participants received standardised exposure to Collaborator's responses, thereby maintaining experimental control and minimising confounding influences on their perceptions. Diego is a robotic head (renamed from its original name by the producer, TIAGo-Head, for ease of distinction~\cite{robotics_tiago_nodate}) with two degrees of freedom in its neck, a screen capable of displaying a face with smooth speaking facial expressions, and a speaker. This Robot Facilitator was placed on the table between the Collaborator and the participant, facing the shared workspace, without physical interaction with it. 

An experimenter was seated next to the Collaborator's emergency button to ensure participant safety, monitoring the system logs of both robots via a laptop, and ready to take over control of the system through a custom-built control interface in case of an error. Participants were informed and clearly understood that the experimenter would not control the robots' behaviour and was present solely as a safeguard for the experiment. 

The Collaborator's computing system ran Ubuntu 20.04 with Robot Operating System (ROS) 1, while the Facilitator's system ran Ubuntu 22.04 with ROS 2. To connect the two robots and the experimenter's laptop, an Ethernet network was set up, with the Collaborator as the server and the others as clients. Signal and data transmission occurred via Ethernet sockets with negligible latency.

Participants were provided with a microphone to capture clear speech when communicating with the Collaborator. An additional laptop was provided for participants to watch the pre-task instruction video and to complete the post-interaction questionnaires. Two tables were used in the room, one for the low-severity task and one for the high-severity task (see~\autoref{fig:exp}). Participants and robots moved between tables when switching tasks to enhance the sense of context change and make the simulation more effective. The task details are explained in the following section.

\subsection{Multi-agentic LLM System and Control Architecture} \label{sec:architecture}

\begin{figure*}
    \centering
    \includegraphics[width=1
\textwidth]{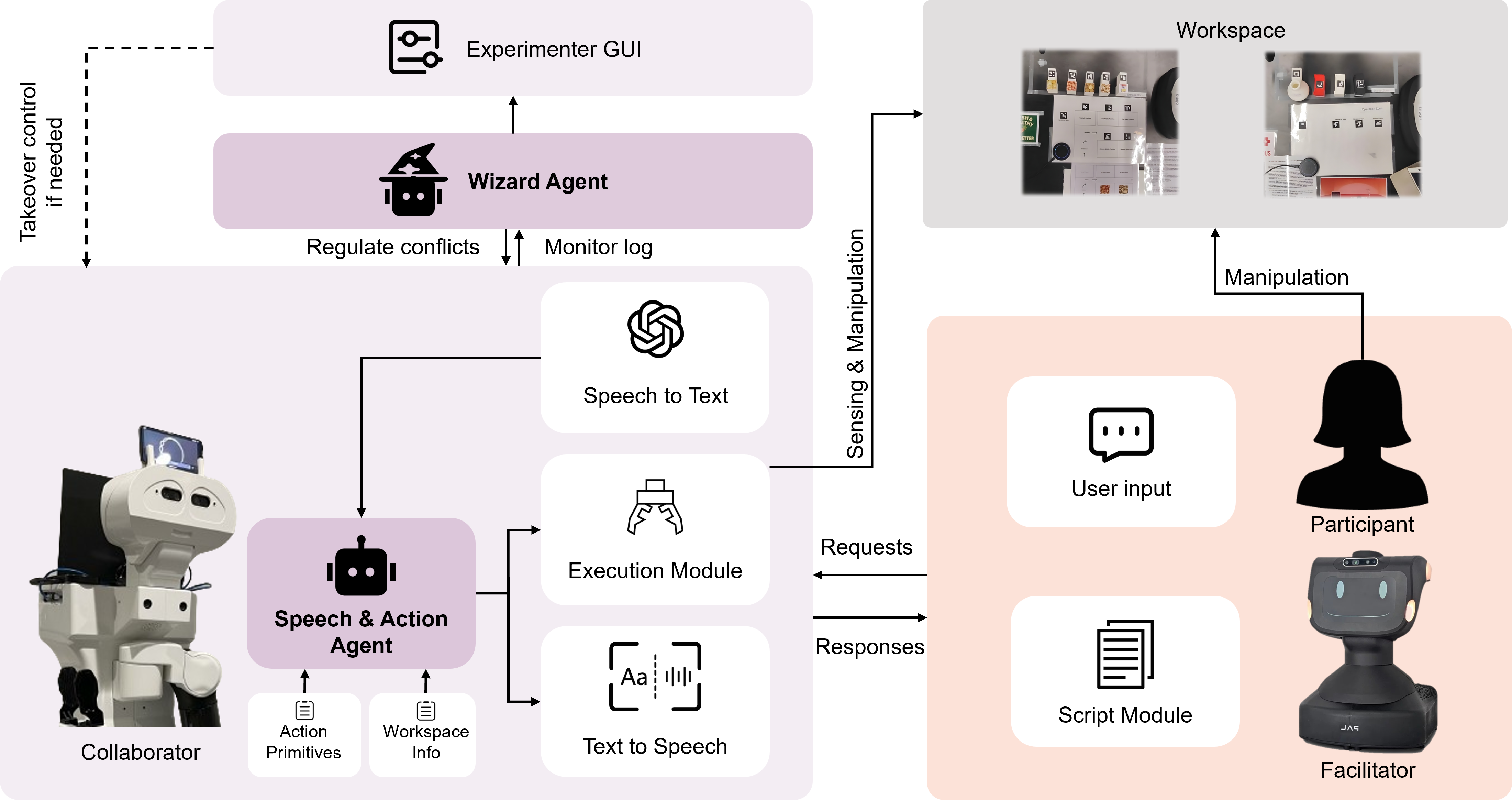}
    \caption{This figure shows the structure of the multi-agentic system. The Collaborator is equipped with a Speech \& Action Agent, which generates verbal responses and action sequences. The Wizard Agent in the Collaborator monitors and regulates system conflicts, and reports them to the Experimenter when takeover control via the GUI is needed. The Facilitator and participants issue requests to the Collaborator, and participants have physical access to the workspace.}
    \Description{This figure shows the system structure. On the left side is the collaborator robot. It connects to a block labelled “Speech and Action Agent.” This agent receives two kinds of information from files: action primitives and workspace information. The agent is linked to three modules: a speech-to-text module, an execution module, and a text-to-speech module. Together, these allow the collaborator robot to understand spoken input, perform physical actions, and generate spoken responses. On top is the wizard agent, which is linked to the speech and action agent to monitor the log and regulate conflicts. On top of the wizard agent is an experimenter GUI, through which the experimenter can take over the control of the robot. On the top right, there is a workspace box showing two small photographs of the task areas: one with grocery items for the low-severity task and one with medical items for the high-severity task. On the right side, the participant is represented by a silhouette connected to a block called user input. Below the participant is the facilitator robot, which has a screen face. This facilitator is connected to a block called the script module. Both the facilitator and participant are connecting with the collaborator to give requests and receive responses. The participant is also connected to the workspace, showing they can manipulate objects.}
    \label{fig:system}
\end{figure*}

To realise the conditions of communication explicitness and intent interpretation correctness in Collaborator's verbal responses, corresponding object manipulation, and interaction between two robots, we designed and implemented a multi-agentic LLM system featuring a graphical user interface (GUI) for the experimenter. The system structure is shown in ~\autoref{fig:system}.

Speech recognition was implemented using OpenAI's real-time transcription model. A microphone streamed audio to the model via API, with voice activity detection identifying the onset and offset of speech. The ``start/stop'' listening signals were transmitted via a wireless connection to the mounted screen, which functioned as an external display and visually indicated the listening status through a web-based interface.

We designed and implemented a mixed-initiative control system based on LLM, consisting of two agents. One Speech \& Action Agent is responsible for generating the verbal responses and actions of the Collaborator according to the verbal input of the participants. Another Wizard Agent provides oversight to the experimenter by monitoring the performance of the system, regulating internal conflicts, and providing summaries and warnings to the experimenter~\cite{schombs_conversation_2026}. The experimenter made decisions on takeover control when it was required. Both agents used OpenAI's GPT-4o model (gpt-4o-2024-11-20) to ensure performance and efficiency. The temperature parameter was set to 0 for both agents to minimise output variability across participants.

The \textbf{Speech \& Action Agent} in the Collaborator was responsible for generating verbal responses and actions based on the experimental condition and participants' real-time requests. Four sets of prompts were tailored for the four conditions. 
The prompts follow the design principles described in~\cite{desmond_exploring_2024}, including the specification of role and persona, brief background, constraints, output format, and few-shot examples.

In the explicit condition, the Speech \& Action Agent always confirmed the inferred intent and action before generating the action sequence by explicitly asking the participant. In the implicit condition, the Speech \& Action Agent never confirmed the intent or action, but instead produced natural verbal responses while generating the corresponding action sequence. Furthermore, in the implicit condition, the Collaborator used the participant's own phrasing for referring to objects or locations in its responses, whereas in the explicit condition, the Collaborator used precise descriptions of objects or locations regardless of the participant's wording.

In the correct condition, the Speech \& Action Agent accurately interpreted the participant's intent, either by explicitly verbalising it (e.g. ``Do you want me to put the test tube in the middle of the table?'') or implicitly by directly executing the action. If the intent was too ambiguous, the agent requested clarification. In the incorrect condition, the Collaborator deliberately misinterpreted the participant's intent in plausible ways, such as by reading the literal meaning of an indirect speech act (e.g. interpreting ``Can you ..?'' as questioning the physical ability ``I have the ability to ...'') or misunderstanding the participant's imperfect but otherwise comprehensible reference to an object. The Collaborator either verbally confirmed the incorrect interpretation or began acting upon it in the implicit condition.

The Speech \& Action Agent accessed two configuration files: one defining action primitives, and the other containing workspace information. The action primitives file listed the Collaborator's atomic actions, which could be integrated into more complex behaviours by the agent to meet participants' requests. The workspace information file contained the information of a list of ArUco markers~\cite{garrido-jurado_automatic_2014}, which are square black-and-white fiducial markers that can be detected by a camera to identify an object and estimate its 3D pose. The file included their IDs, and names and simple descriptions of the objects or locations they represented. By using OpenCV and ArUco markers, the Collaborator maintained semantic information and 3D poses of target objects and locations more efficiently, accurately, and with less processing time compared to visual scene understanding or multimodal LLM agents. Since the focus of this study was not on developing a state-of-the-art multi-agentic collaborative robot system, we opted for action primitives and ArUco markers to minimise technical issues.

Another LLM agent, the \textbf{Wizard Agent}, continuously ran in the background, subscribing to all system topics, including participants' input utterances, the outputs of the Speech \& Action Agent, task progress, and the status of each module. When an error or warning occurred, the Wizard Agent generated a concise, readable summary for the experimenter, who then decided whether to take over control from the agents. The Wizard Agent also regulated the sequence of processes and any signal conflicts, including speech-to-text, LLM processing, text-to-speech, the Collaborator's screen display, action execution, quick error recovery, and scripted interaction with the Facilitator robot. Because the Wizard Agent was itself LLM-based, its summaries could in principle contain errors or hallucinations. Its role was therefore deliberately restricted to advisory reporting, and it had no authority to alter the Collaborator's behaviour on its own. All control decisions remained with the experimenter, who could inspect the raw system logs alongside the Wizard Agent's summaries through the GUI.

The interaction between the two robots was script-based to ensure consistency across participants. The Collaborator and Facilitator both used the text-to-speech API provided by their manufacturer, and each employed a distinct natural-sounding voice to remain easily distinguishable. Their communication was observable to participants through verbal communication as well as head and gaze movements, while signals were exchanged internally via Ethernet sockets to ensure stable communication across the different ROS systems.

\subsection{Task Design}
In this study, two collaborative tasks were designed: a low-severity task simulating the stocking of items in a grocery store, and a high-severity task simulating the preparation of a medical injection for a critically ill patient. The goal of these tasks was to simulate different levels of situational risk~\cite{xie_risk_2003}, giving participants a sense of severity without causing any physical or psychological harm. The design followed the principle of varying along three dimensions:
\begin{itemize}
    \item Presence or absence of ingestible or bodily-administered elements
    \item Involvement or non-involvement of vulnerable groups
    \item Presence or absence of severe consequences if performed incorrectly
\end{itemize}

In the \textbf{stocking task}, participants were told that they were working in a grocery store, responsible for stocking food from the storage room onto the display shelf (see the right of~\autoref{fig:exp}(b)). Only the Collaborator had access to the storage room and could retrieve items. Participants were given an incomplete floor plan showing the items that should be placed on the bottom row, as well as the positions of the entrance, hallway, and exit. They were also informed that the Facilitator would be responsible for the items placed on the top row. The food items were represented by 3D-printed props with handles for robot grasping. Printed food photos were attached to the front of the props. The simulated store shelves were represented by a printed floor layout placed on the workspace.

In the \textbf{medical task}, participants were told to imagine they were working in a hospital laboratory, and were required to prepare a medical injection solution for a critically ill patient (see the left of~\autoref{fig:exp}(b)). The props included 3D-printed test tubes containing liquid with an unstable base, and three 3D-printed boxes of different colours and containing different amounts of powder. The props were initially placed in a sterile zone, and participants needed to ask the Collaborator to move the ones they needed into the shared workspace, placing each of them at one of four preferred spots. Participants were then responsible for pouring the powder into the test tube, while the Facilitator was responsible for delivering instructions for the subsequent steps. A timer was placed beside the workspace, set to count down from ten minutes. The timer was intended to convey a sense of severity~\cite{kontogiorgos_behavioural_2020}, even though participants were never directly instructed to complete the task within the time limit. The high severity of this task was primarily situational~\cite{xie_risk_2003}, reflecting the potential for severe consequences if such a procedure were performed incorrectly.

The two tasks were set up at different tables, requiring participants and robots to move between them when switching tasks to indicate a clear transition of the context. Decorative pictures matching each scene were placed beside the workspaces to enhance the sense of scenario immersion. In both tasks, participants were instructed that they could communicate with the Collaborator in any way they felt natural and comfortable, both in giving instructions and in referring to objects and positions. Participants were informed that the Collaborator did not have access to task steps and would rely on instructions provided either by them or by the Facilitator. Participants shared physical abilities with the Collaborator, who could manipulate objects they could not reach, and shared partial task information with the Facilitator, who knew steps they did not. As a result, participants had to collaborate physically with the Collaborator, while depending on the Facilitator to provide certain instructions to the Collaborator. Each task began with a warm-up step (see~\autoref{fig:exp}(b)), in which participants asked the Collaborator for either a floorplan documentation number or a patient ID from its internal system. This initial verbal-only exchange helped participants become familiar with the Collaborator before beginning the main actionable steps.

\subsection{Scripted Interaction between Robots}
The Facilitator was included in the experimental design to ensure that each participant experienced at least three instances of the Collaborator responding to indirect requests under the assigned condition. This design allowed participants both to experience the Collaborator's responses to their own requests during collaboration and to observe the Collaborator's responses to designed indirect requests. The interaction between the two robots was scripted, as LLM outputs can vary slightly across runs, whereas this study required controlling the consistency of experience across participants. ~\autoref{apd:example} shows an example of the Facilitator's indirect requests and the Collaborator's responses under different conditions. The indirect requests were initially developed based on non-conventionalised indirect requests~\cite{zhang_can_2025}, which require interpretation through conversational context, physical environmental context, and pragmatic features. The final utterances were refined through several iterations, informed by the results of nine pilot sessions. These indirect requests were phrased to be as natural as human speech while avoiding excessive vagueness that could cause misinterpretation, a balance that was further validated through participants' interview data.

\subsection{Data Collection and Analysis}

\subsubsection{Quantitative Data}

We collected quantitative data using a questionnaire (see~\autoref{apd:questionnaire}) and qualitative data through a semi-structured interview. The following dependent variables were collected:
\begin{itemize}
    \item Perceived Grounding with the Collaborator: To examine the effects of communicative explicitness and interpretation correctness, participants were asked to rate their perception of grounding with the Collaborator three times per task. Perceived grounding was measured using the 5-point Likert scale~\cite{joshi_likert_2015}. The timing of the ratings was treated as a random effect in the statistical model to account for the repeated-measures design.
    \item Human Communication Effort: At the end of each task, participants rated the amount of effort they put into communicating with the Collaborator. This was measured on a 5-point Likert scale.
    \item Collaborator Communication Effort: At the end of each task, participants also rated their perception of how much effort the Collaborator put into communication during the collaboration. This was measured on a 5-point Likert scale.
\end{itemize}

To analyse the effects of Explicitness and Correctness in the Collaborator's responses on grounding and communication effort, as well as the moderating effect of task severity and the covariate of participants' prior experience with physical collaborative tasks, we used Cumulative Link Mixed Models (CLMMs) implemented via the \textit{ordinal} package in R~\cite{christensen_ordinal_2010}. CLMMs were chosen because they respect the ordinal nature of Likert-type ratings rather than assuming equal distances between response categories, and they allow the between-subject factors, the within-subject moderator, and the covariate to be tested within a single model rather than through separate tests. This single-model approach also avoids the metric assumptions imposed by ANOVA-based analyses and the reduced sensitivity of rank-based non-parametric tests, both of which remain common for ordinal data in human-computer interaction (HCI) research despite their limitations~\cite{victor_syiem_better_2026}. Estimated marginal means (EMMs) were computed using the \textit{emmeans} package with Holm correction applied~\cite{searle1980population}. Two model structures were fitted, with the effort structure applied separately to human and Collaborator communication effort:
{
\begin{align}
\text{Grounding} \sim{}& \text{Explicitness} \times \text{Correctness}
  + \text{Correctness} \times \text{Severity} + \text{Explicitness} \times \text{Severity} \nonumber\\
&+ \text{Prev.collab.} + (1+\text{timing}\mid\text{pid}) + (1\mid\text{timing})
  \label{eq:clmm-grounding}\\[0.5em]
\text{Effort} \sim{}& \text{Explicitness} \times \text{Correctness}
  + \text{Correctness} \times \text{Severity} + \text{Explicitness} \times \text{Severity}  \nonumber\\
&+ \text{Prev.collab.}
  + (1\mid\text{pid}) \label{eq:clmm-effort}
\end{align}
}
In both models, $\times$ denotes an interaction together with its constituent main effects, so the specifications test the main effects of $Explicitness$ and $Correctness$, their interaction, and the moderating effect of $Severity$ on each. $Prev.collab.$ is a covariate capturing participants' prior experience with physical collaborative tasks. Terms in parentheses are random effects, comprising a by-participant intercept ($pid$) in both models, and, for perceived grounding, additional terms accounting for the three rating $timing$s within each task.
Effect sizes were reported as odds ratios (ORs)~\footnote{For example, an $OR = 3$ means that a one-unit increase in the predictor makes the outcome variable three times more likely to fall in the other category of the outcome.}, which are appropriate for CLMMs and allow interpretable comparisons of predictors while respecting the ordinal nature of ratings~\cite{ananth_regression_1997}.

\subsubsection{Qualitative Data}

At the end of the experiment, we conducted a semi-structured interview in English, lasting approximately 20 minutes to gather qualitative feedback from participants. Interviews were audio-recorded and transcribed by the first author. The interviews were intended to supplement the quantitative results and provide insight into participants' subjective experiences of the collaboration. We began by asking whether participants noticed the specific ways in which the Collaborator communicated, the naturalness of the Facilitator's indirect requests, and the differences between the task contexts, in order to evaluate the validity of the condition manipulations. Given the between-subjects design of the study, we explained the experimental conditions that participants had not experienced, ensuring they left with a comprehensive understanding of the study.

The interview results were analysed using reflexive thematic analysis (RTA), which was well suited to this study because it emphasises the researcher's active role in constructing themes, thereby fostering flexibility, creativity and critical reflection. Moreover, the aim of the study is to explain how participants experienced the manipulated differences in the Collaborator's behaviour. We chose reflexive over codebook approaches because the analysis was intended to develop an explanation of why the observed effects took the form they did, which required interpretive latitude that a fixed coding frame would have constrained. This approach permits the integration of the researcher's insights and observations from the experimental process, making it particularly effective for exploring subtle phenomena~\cite{braun_doing_2022}.

Following the 6-phase guidance by~\cite{braun_using_2006}, the first author of this paper, who possesses substantial expertise in human-robot collaboration and communication, conducted the RTA, and had iterative discussions with the research team to further reflect on and enrich the analysis. In phase 1, the researcher thoroughly reviewed all transcripts. In phase 2, the researcher inductively generated initial codes at the sentence level, which were either semantic, representing participants' explicit accounts, or latent, reflecting deeper meanings inferred from the data in light of the researcher's knowledge background. In phase 3, the researcher constructed the initial themes and categorised the codes. In phase 4, the whole research team reviewed the themes through multiple rounds. In phase 5, the themes were defined and named. In phase 6, the team drafted the initial report of the qualitative analysis. Phases 4 to 6 were repeated over several rounds, during which the themes were iteratively refined. This process aligns with the principles of RTA, which emphasise continuous reflexivity and recursive engagement with the data~\cite{terry_themantic_2017}.

\subsection{Participants and Procedure}
\begin{figure}
    \centering
    \includegraphics[width=1\linewidth]{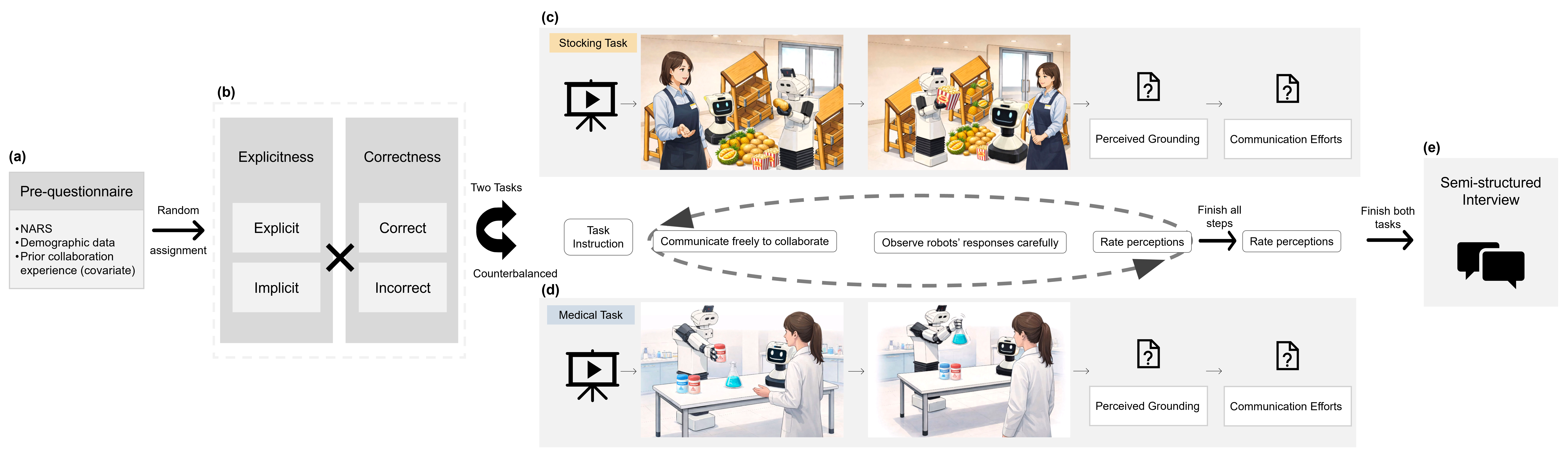}
    \caption{This figure illustrates the experiment procedure. Each participant completed a pre-questionnaire (a) before being randomly assigned to one of the four conditions (b). The participants then performed the low (c) and high (d) severity tasks with Collaborator and Facilitator in a counterbalanced order. Before each task, participants watched an instruction video and read a task description. After communicating freely with the Collaborator and observing its responses to indirect requests after each step, participants were asked to rate their perceptions of grounding. Participants were asked to rate their perceptions of the communication efforts at the end of each task. After finishing two tasks, the experiment concluded with a semi-structured interview (e).}
    \label{fig:procedure}
\end{figure}

An a priori power analysis was conducted using \textit{G*Power}~\cite{faul_statistical_2009, faul_gpower_2007} to obtain an initial estimate of the required sample size. Because no closed-form power calculation exists for Cumulative Link Mixed Models, the calculation was necessarily an approximation based on linear multiple regression (fixed model, single regression coefficient)~\cite{kumle_estimating_2021}, which is a common approach in behavioural and HCI research. The analysis parameters were adopted from closely related work~\cite{zhang_can_2025}, with an effect size of 0.25 (medium to large), an alpha level of 0.05, a desired power of 0.8, 7 predictors, and a one-tailed test. This yielded a required sample of 27 participants, and a simulation-based power analysis using \textit{simr}~\cite{green_simr_2016} produced a similar estimate (N = 28). The final sample size was rounded up to 32, in line with reported norms for mixed-design experiments in HCI research~\cite{caine_local_2016} and HRI research using CLMM analysis and \textit{G*Power}~\cite{bilal_design_2026, tabatabaei_oops_2026}. The study was exploratory in nature, and the model estimates were complemented by a qualitative analysis of interviews with all participants, providing a convergent source of support alongside the statistical models. Moreover, CLMMs leverage both the repeated measures and the ordinal structure of the data, offering greater statistical sensitivity at a given sample size than traditional non-parametric alternatives~\cite{victor_syiem_better_2026}.

To assess attitudes toward robots, we employed the interaction subscale of the Negative Attitude Toward Robots Scale (NARS)~\cite{nomura_measurement_2006} before the experiment, selecting three questions pertinent to engagement in conversation and collaboration with a robot (see \autoref{apd:NARS}). These items were administered as a screening tool to exclude participants with pronounced negative dispositions towards robots, who might otherwise experience discomfort during the study. Each session lasted approximately 60 minutes, and participants received an AUD16 voucher as compensation. Ethical approval for the study was granted by the Institutional Review Board (IRB).

As a result, we recruited 32 participants (19 females and 13 males) through the university online notice board and street flyers. Participants were aged between 18 and 38 ($M = 26.3, SD = 5.4$), and none showed strongly negative attitudes towards robots. 81.25\% of them had prior experience of doing physical collaborative tasks with either human or robot teammates. The details of participants' demographic information are shown in~\autoref{tb:demo}. All participants received an information sheet and signed a written consent form prior to participation.

\begin{table*}[ht]
\caption{Overview of participants' demographic information and their prior experience of doing physical collaborative tasks. (Rarely: less than once a month; Sometimes: at least once a month but less than once a week; Often: at least once a week but less than once a day; Very often: at least once a day.)}
\label{tb:demo}
\begin{tabular}{lclclc}
\hline
\multicolumn{2}{c}{Gender} & \multicolumn{2}{c}{Age} & \multicolumn{2}{c}{Physical Collaborative Tasks} \\ \hline
Female & 59.38\% & 18-25 & 50.00\% & Never      & 18.75\% \\
Male   & 40.63\% & 26-35 & 40.63\% & Rarely     & 15.63\% \\
       &        & 36-45 & 9.38\%  & Sometimes & 31.25\% \\
       &        &       &        & Often      & 31.25\% \\
       &        &       &        & Very often & 3.13\%  \\\hline
\end{tabular}
\end{table*}

The experimental procedure is shown in ~\autoref{fig:procedure}. 
After welcoming the participants and providing safety instructions for interacting with the robots, the study began with a pre-questionnaire that captured participant demographics and the extent of their prior experience in performing physical collaborative tasks. Prior experience was treated as a covariate in the data analysis, as it has been shown to influence participants' perception in HRC~\cite{zhang_can_2025}. Each participant interacted under one of the four conditions (IC, II, EC, or EI, see \autoref{tab:conditions}) and completed both tasks in a counterbalanced order. Before each task, participants were provided with a video introducing the task context (i.e. grocery store or hospital laboratory) and a written description outlining the steps involved. In addition, the two robots gave a brief introduction of themselves to familiarise participants with their voices. 

Participants were required to lead the collaboration with the Facilitator and verbally relay the steps and objectives to the Collaborator. After participants completed their own steps, they were asked to observe the Facilitator giving instructions to the Collaborator and to complete the other half of the task. Before the Collaborator responded or performed any action, participants were asked to interpret what the Facilitator was trying to instruct the Collaborator to do. They then rated their sense of grounding with the Collaborator, via a questionnaire, once the Collaborator had completed the action. After each task, participants were also asked to rate their perception of their own communication effort when interacting with the Collaborator, as well as their perception of the Collaborator communication effort. Finally, the study concluded with a semi-structured interview.

\subsection{Validation Check}
Before the start of the interview, we asked participants about their feelings about the task itself and the indirect requests from the Facilitator.

As a result, most participants (30 out of 32) mentioned that they noticed a difference in the severity of the tasks, and 2 did not answer this question explicitly. Participants noted that the medical task was of high severity because it involved a life-saving situation. Some mentioned that the sense of severity and tension came from the timer. A few thought that the precision required in the medical operation made the task more severe. $P2_{EI}$ believed the grocery task was more ordinary and simpler. However, 6 of them also remarked that although they recognised the difference in stakes, it did not affect their feelings because they knew it was only a simulation conducted in a laboratory experiment.

All participants recognised that the Facilitator intentionally used indirect requests. Twenty-seven found these requests natural and intuitive. One reported that the issue was with the clarity of the voice rather than the utterance itself, and four mentioned a case they found a bit ambiguous. The Facilitator's speech was described as human-like, relatable, and straightforward. ``Most of the time in daily life, I would say exactly the same words as Diego did, and other people would understand me'' said $P6_{EI}$. 

\section{Quantitative Findings}
In the following sections, we present the key results of the CLMM analysis. The key results are shown in~\autoref{tab:clmm_results} and full results are shown in~\autoref{tab:clmm_all}.

\begin{table}[t]
\centering
\caption{Key results for CLMM analysis}
\label{tab:clmm_results}
\begin{tabular}{lrrrr}
\hline
Predictor & Estimates ($\beta$) & Effect Sizes (OR) & Std. Error & p-value\\
\hline
\multicolumn{4}{l}{\textit{Grounding with Collaborator}}\\

Explicit                    &  2.677  & 14.54      & 1.610 & 0.096. \\
Incorrect                   & -10.737 & 46027.77   & 1.950 & \textless{}0.001* \\
High sev.                   & -0.434  & 1.54       & 0.838 & 0.604 \\
Prev.collab.               &  0.302  & 1.35       & 0.378 & 0.425 \\
Explicit $\times$ High      & -1.334  & 3.80       &0.925 & 0.149 \\
Incorrect $\times$ High     &  0.967  & 2.63       &1.006 & 0.337 \\
Explicit $\times$ Incorrect & -2.000  & 7.39      &1.708 & 0.242 \\
\hline

\multicolumn{4}{l}{\textit{Human Communication Effort}}\\

Explicit                    &  1.200 & 3.32       & 0.002 & \textless{}0.001* \\
Incorrect                   &  3.910 & 49.90      & 1.623 & 0.016* \\
High sev.                   &  0.269 & 1.31      & 0.002 & \textless{}0.001* \\
Prev.collab.               & -0.112 & 1.12       & 0.002 & \textless{}0.001* \\
Explicit $\times$ High      & -0.249 & 1.28      & 0.002 & \textless{}0.001* \\
Incorrect $\times$ High     &  0.376 & 1.46      & 0.805 & 0.640 \\
Explicit $\times$ Incorrect & -3.632 & 37.79       & 1.940 & 0.061. \\
\hline

\multicolumn{4}{l}{\textit{Collaborator Communication Effort}}\\

Explicit                    & -0.108 & 1.11       & 1.440 & 0.940 \\
Incorrect                   & -2.322 & 10.20     & 1.481 & 0.117 \\
High sev.                   &  1.776 & 5.91      & 0.932 & 0.057. \\
Prev.collab.               &  0.123 & 1.13       & 0.418 & 0.768 \\
Explicit $\times$ High      & -1.735 & 5.67       & 1.046 & 0.097. \\
Incorrect $\times$ High     & -1.702 & 5.49      & 1.047 & 0.104 \\
Explicit $\times$ Incorrect &  4.728 & 113.07     & 2.050 & 0.021* \\
\hline
\end{tabular}
\end{table}

\subsection{Grounding via Communication}
The correctness of the Collaborator's contextual intent interpretation was the only predictor significantly associated with perceived grounding. When the Collaborator's interpretations were incorrect, participants reported lower grounding ($\beta = -10.737$, $SE = 1.950$, $p <0.001$) (RQ1.1). The magnitude of this estimate reflects a limited overlap between correct and incorrect conditions in the grounding rating, consistent with the pattern observed in~\autoref{fig:clmm_plot}. No other main effects, interactions, or covariates reached significance (RQ1.2, RQ2.1, RQ2.2, RQ3). In this experiment, correctness of intent interpretation was thus the primary factor associated with participants' sense of common ground with the robot.

\subsection{Human Communication Effort}
For human communication effort, participants reported higher ratings when the Collaborator responded explicitly ($\beta = 1.200$, $SE = 0.002$, $p<0.001$) (RQ2.1) and when its interpretations were incorrect ($\beta = 3.910$, $SE = 1.623$, $p = 0.016$) (RQ1.1). Task severity also showed a positive association with reported effort ($\beta = 0.269$, $SE = 0.002$, $p <0.001$). Regarding its moderating role, severity attenuated the effect of explicitness ($\beta = -0.249$, $SE = 0.002$, $p <0.001$), suggesting that the additional effort associated with explicit responses was somewhat smaller under high-severity tasks (RQ3). Severity did not moderate the effect of correctness. The explicitness-correctness interaction was not significant (RQ1.2, RQ2.2). Finally, prior experience with physical collaborative tasks was associated with lower reported effort ($\beta = -0.112$, $SE = 0.002$, $p <0.001$).

\subsection{Robot Collaborator Communication Effort}
For perceptions of the Collaborator communication effort, neither explicitness nor correctness showed a significant main effect, and severity did not significantly moderate these effects (RQ1.1, RQ2.1, RQ3). The interaction between explicitness and correctness was significant ($\beta = 4.728$, $SE = 2.050$, $p = 0.021$). Participants rated the Collaborator's effort higher when explicit responses were paired with incorrect interpretations (RQ1.2, RQ2.2). Given that this is a single interaction effect estimated with a moderate standard error, we interpret it cautiously as suggestive rather than conclusive in quantitative findings.

In summary, participants' perceptions of their own effort were associated with correctness, explicitness, and severity, whereas their perceptions of the robot's effort were associated only with the combination of explicitness and correctness. These patterns offer a preliminary picture of how communication effort may be perceived and distributed in HRC, which we examine further in the qualitative findings.

\begin{figure*}
\centering
\includegraphics[width=1\textwidth]{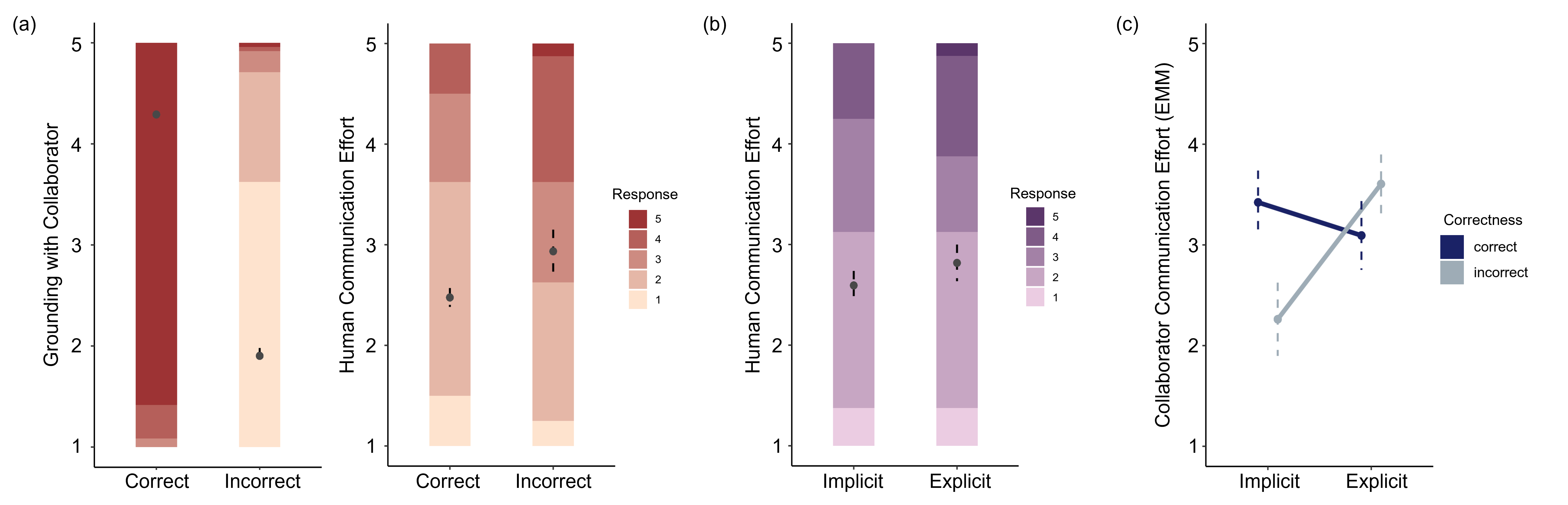}
\caption{(a) Stacked bar chart with estimated marginal means (EMMs) overlay for $Correctness$'s significant effects. (b) Stacked bar chart with EMMs overlay for $Explicitness$'s significant effects. (c) The plot for the $Correctness$ and $Explicitness$ interaction effect.}
\Description{This figure presents quantitative results from the experiment in three panels. Panel a shows stacked bar charts with estimated marginal means overlaid for the effect of correctness. The left chart measures grounding with the collaborator, and the right chart measures human communication effort. For grounding, ratings are higher when responses are correct and lower when responses are incorrect. For human communication effort, the effort is rated higher when responses are incorrect. Panel b shows stacked bar charts with estimated marginal means overlaid for the effect of explicitness on human communication effort. Effort is rated lower in the implicit condition and higher in the explicit condition. Panel c shows a line plot of the interaction effect between correctness and explicitness for collaborator communication effort. Two lines are plotted: one for correct responses and one for incorrect responses. In the implicit condition, correct responses are associated with higher collaborator effort than incorrect ones. In the explicit condition, the pattern reverses, with incorrect responses associated with higher collaborator effort. Error bars are shown across all plots.}
\label{fig:clmm_plot}
\end{figure*}

\section{Qualitative Findings}
\subsection{Corpus and Analytic Scope}
The corpus comprised 32 interviews, eight per condition. We constructed three major themes, defined in~\autoref{tab:themes}. The table lists the participants whose extracts contributed to each theme. These lists support traceability and are not prevalence estimates as RTA does not treat the number of participants articulating a pattern as an index of its analytic importance~\cite{braun_one_2021, maxwell_using_2010}. Three constraints on interpretation apply throughout. First, our analysis concerns how participants made sense of the collaboration and communication during tasks and any reported shifts in their own way of speaking are self-reported. Second, participants were interviewed after both tasks, so descriptions of the two severity levels were produced retrospectively. Third, accounts of conditions participants did not experience are comparative judgments formed after those conditions were described to them. Participant identifiers carry condition as a subscript ($E$/$I$ for explicit/implicit responses, $C$/$I$ for correct/incorrect interpretation).

\begin{table*}[t]
\caption{Theme definitions, boundaries, and contributing participants from the reflexive thematic analysis of post-study interviews.}
\label{tab:themes}
\begin{tabular}{p{0.17\linewidth} p{0.245\linewidth} p{0.245\linewidth} p{0.25\linewidth}}
\hline
\textbf{Theme} & \textbf{Description} & \textbf{Boundary} & \textbf{Contributing Participants} \\
\hline
1. Common ground is perceived through the robot's uptake and is continually negotiated
& Captures how participants judged whether common ground was held, on the basis of the Collaborator's interpretive accuracy, and the three relations they described, including stable, repeatedly repaired, or absent but worked around.
& Includes comments on whether and how mutual understanding was attained. Excludes comments on the effort this required, which fall under Theme 2, and the influence of risk, which fall under Theme 3.
& $P1_{IC}$, $P5_{IC}$, $P9_{IC}$, $P11_{IC}$, $P17_{IC}$, $P32_{IC}$; 
$P4_{EC}$, $P8_{EC}$, $P14_{EC}$, $P20_{EC}$, $P23_{EC}$, $P29_{EC}$; 
$P3_{II}$, $P7_{II}$, $P13_{II}$, $P15_{II}$, $P19_{II}$, $P27_{II}$, $P28_{II}$, $P30_{II}$;
$P2_{EI}$, $P6_{EI}$, $P12_{EI}$, $P16_{EI}$, $P22_{EI}$
\\
\addlinespace
2. Effort is perceived through who carries it and whether it feels needed
& Captures how participants attributed the effort of communication to themselves, the Collaborator, or both, and how those attributions formed the basis of their evaluations of the collaboration.
& Four sub-themes correspond to the distributions participants described. Excludes how severity reframes the meaning of extra effort, which falls under Theme 3.
& $P1_{IC}$, $P5_{IC}$, $P9_{IC}$, $P11_{IC}$, $P17_{IC}$, $P32_{IC}$; 
$P4_{EC}$, $P8_{EC}$, $P14_{EC}$, $P18_{EC}$, $P20_{EC}$, $P23_{EC}$, $P26_{EC}$, $P29_{EC}$; 
$P3_{II}$, $P7_{II}$, $P13_{II}$, $P15_{II}$, $P19_{II}$, $P27_{II}$, $P28_{II}$;
$P2_{EI}$, $P6_{EI}$, $P10_{EI}$, $P12_{EI}$, $P16_{EI}$, $P22_{EI}$, $P25_{EI}$, $P31_{EI}$\\
\addlinespace
3. Task severity frames the meaning of extra effort in collaboration
& Captures how participants' evaluations of the same communicative behaviour shifted with the severity of error consequences, positioning severity as a backdrop that cuts across Themes 1 and 2.
& Includes severity effects on expectations, referring expressions, request directness, and pace of interaction. Excludes effects on task performance, and accounts where severity is not invoked.
& $P1_{IC}$, $P11_{IC}$, $P17_{IC}$, $P21_{IC}$, $P32_{IC}$; 
$P4_{EC}$, $P8_{EC}$, $P14_{EC}$, $P20_{EC}$, $P26_{EC}$, $P29_{EC}$; 
$P3_{II}$, $P7_{II}$, $P13_{II}$, $P15_{II}$, $P19_{II}$, $P27_{II}$, $P28_{II}$;
$P2_{EI}$, $P6_{EI}$, $P10_{EI}$, $P16_{EI}$, $P31_{EI}$ \\
\hline
\end{tabular}
\end{table*}

\subsection{Common Ground Is Perceived through the Robot's Uptake and Is Continually Negotiated}
While participants' perception of grounding was primarily shaped by the Collaborator's ability to interpret their intent accurately, three distinct situations emerged: grounding could be sustained when mutual understanding was achieved, compromised when only partial alignment was established, or entirely forgone when no common ground was formed, yet collaboration was still realised through alternative means.

In the $Explicit \times Correct$ and $Implicit \times Correct$ conditions, participants reported experiencing mutual understanding with the Collaborator. The common ground was perceived as stable, requiring no re-establishment and enabling smooth collaboration (e.g. $P1_{IC}$, $P5_{IC}$, $P9_{IC}$). For $P14_{EC}$, confirmation was incidental to grounding, since understanding was felt to be in place whether or not the Collaborator said so. $P8_{EC}$ remarked that grounding was most salient in the Collaborator's uptake of a distinctive spatial description. These point to an outcome-based criterion that intent was treated as grounded if the task was completed, irrespective of whether understanding had been verbally confirmed along the way.

\begin{quote}
$P8_{EC}$: ``That (grounding with Tiago) was very good. When I said, `Can you place the potato near the exit?', he inferred (as) put it down on the bottom (shelf). ... I feel like he's (Tiago) quite good, because I thought maybe he would put it in the middle (of the hallway) and I'm going to move it (by myself).''
\end{quote}

In the $Explicit \times Incorrect$ and $Implicit \times Incorrect$ conditions, grounding was more fragile. Several participants noted that they were unable to establish grounding immediately. For instance, $P7_{II}$ and $P13_{II}$ explained that mutual understanding only emerged from the second round of communication, once they provided more specific requests. $P2_{EI}$ believed that they achieved only moderate grounding with the Collaborator, as this was limited to direct requests. Minor misreadings were treated as sufficient to unsettle the whole, so grounding was continually rebuilt through rephrasing and repeated attempts.

Some participants reported never achieving common ground. Beyond noting that the Collaborator did not understand their intents, participants also expressed doubts and confusion about why the Collaborator was unable to interpret their requests. $P19_{II}$ observed that the Collaborator appeared to understand ``what'' action to take but seemed never to grasp the underlying ``why''. Similarly, $P27_{II}$ and $P28_{II}$ remarked that they believed the Collaborator failed to comprehend the broader context of the task and instead focused only on small, isolated details. 
Nevertheless, the Collaborator's misunderstandings were perceived differently across participants.

When misunderstandings were resolved in the second round of communication and occurred in the low-severity task, they were perceived as acceptable and tolerable. $P28_{II}$ accepted the Collaborator's mistakes, remarking, ``That's how it is''. $P13_{II}$ observed that even in the $Implicit$ condition, when the Collaborator began executing an incorrect action without confirming, there was still at least an opportunity for them to correct it. 

\subsection{Effort Is Perceived through Who Carries It and Whether It Feels Needed}
\subsubsection{\textbf{Least Communication Effort}}
Participants who reported perfect grounding with the Collaborator felt that collaboration required very little effort, since the Collaborator appeared to understand whatever they said. Once established, grounding licensed a shift towards indirectness. Participants attributed contextual awareness to the Collaborator and spent that attribution by asking less explicitly for what they wanted ($P5_{IC}$, $P8_{EC}$). The Collaborator's uptake performance also shaped participants' broader perceptions of the robot. Participants in the $Correct$ conditions primarily offered positive evaluations, describing the Collaborator as human-like, intelligent, cooperative, and skilled. As $P11_{IC}$ remarked, ``I feel like Tiago actually gave more of a human-like way of talking.''

Participants converged on the same word ``straightforward'' to describe opposite strategies, and the divergence is instructive. For one group, straightforwardness meant indirectness, by saying whatever came to mind without pre-formulating it, which was experienced as lower in cognitive cost and closer to ordinary talk ($P4_{EC}$, $P11_{IC}$, $P29_{EC}$), as well as more comfortable and more nuanced ($P4_{EC}$, $P5_{IC}$). Initial specificity was abandoned once communication ran smoothly ($P1_{IC}$, $P9_{IC}$). For another group, straightforwardness meant directness, which was an economy of phrasing, familiar from ChatGPT or Siri ($P26_{EC}$), and chosen precisely because it was legible to machines ($P3_{II}$, $P28_{II}$). $P8_{EC}$, a K-12 teacher, drew an analogy to instructing children. The two groups share the goal of minimising their effort but differ in whether they understand that effort to reside in formulating an utterance or in ensuring it lands.
Notably, participants in the $Implicit \times Correct$ condition generally adopted a more natural, indirect style, offering nuanced ways of speaking. They also became increasingly comfortable and relaxed, resembling how they would normally interact with another human.

\subsubsection{\textbf{Burden on the Human}}
Participants reported several situations that demanded greater effort in communication and collaboration. Misinterpretation redirected effort towards simplification and pre-emption, with phrasing monitored in advance for the misreadings it might invite. A second and less visible cost lay in detection, as participants had to infer from ambiguous responses and actions whether they had been understood at all, work they described as both demanding and tiring. In the $Implicit \times Incorrect$ condition, the cost was registered not as difficulty but as friction. Communication required a pause for planning, which participants contrasted with the fluency of talking to a person ($P19_{II}$, $P27_{II}$). As $P15_{II}$ put it, ``You can't make assumptions that it will just understand.'' Participants searched for the correct keyword, treating comprehension as lexical retrieval rather than inference ($P4_{EC}$, $P9_{IC}$, $P14_{EC}$). But it was costly to act on, and as
$P6_{EI}$ noted, ``I'm repeating the same thing but in a different way.'' Participants described feelings of frustration and fatigue when giving direct requests. $P19_{II}$ concluded, ``I found towards the end just getting a bit resentful. ... I don't want Tiago to be my colleague.''

Moreover, when the grounding broke down, re-establishing it required additional effort. $P19_{II}$ reported having to ``think the way Tiago thinks'' suggesting that the burden of maintaining mutual understanding fell more heavily on participants. In the $Implicit \times Incorrect$ condition, participants were unsettled by the Collaborator's lack of confirmation and incorrect actions, which initially made them question themselves, then left them frustrated, and ultimately prompted them to shift towards direct requests or even to devise their own opportunities for confirmation. For example, $P28_{II}$ created opportunities for explicit confirmation to preempt misunderstanding in the high-severity task, which further added to their workload.

\begin{quote}
$P28_{II}$: `` I first confirmed whether it (Tiago) knew where was the 30 grams of powder. Then it confirmed that it was the white one, so I knew it would pick the right one. I'm trying to avoid picking the wrong one and put more (effort) in another round (of communication).''
\end{quote}

Regardless of whether the Collaborator interpreted correctly, participants emphasised that responding to each of the Collaborator's confirming questions created additional effort. $P6_{EI}$ perceived the Collaborator's confirmation of a wrong intent as minimal effort, while requiring extra effort from humans. They viewed the Collaborator as an order-following assistant. ``(Tiago was like:) That's what I'm doing, and hope for the best. Correct me if I'm wrong.'', said $P19_{II}$.

However, mistakes were occasionally seen as making the Collaborator more human-like and relatable, but not more intelligent. For some, the Collaborator felt like a teammate, albeit a terrible one. Interestingly, $P27_{II}$ remarked, ``Whereas I think if he (Tiago) got everything right, I may not have felt as empathetic.''

Surprisingly, in the $Explicit \times Correct$ condition, participants reported more effort of their own, although the Collaborator understood every intent and they saw it as investing effort. This effort arose not only from responding to the robot's confirmations but also from merely observing it doing so, especially in the low-severity task. Participants felt that the Collaborator did not need to respond in detail every time, and that its extra effort left them confused and impatient ($P18_{EC}$, $P23_{EC}$, $P26_{EC}$). $P20_{EC}$ chose to front-load effort by crafting precise, direct instructions to avoid confirmation, even though it was correct. As $P23_{EC}$ explained, while confirmation is sometimes very necessary, there were moments when they ``just wanted to press a skip button,'' wishing to hear confirmations only when a misinterpretation was foreseeable. Some described the Collaborator as annoying and confusing, and as machine-like in a way they did not welcome in this situation.

\subsubsection{\textbf{Burden on the Robot}}
In some conditions, participants also perceived communication effort on the robot's side. Participants recognised that their indirectness shifted effort to the listener ($P20_{EC}$, $P26_{EC}$), here, the Collaborator. ``It's (Tiago) like a teammate, but more like, processing what we say, like an exhausted one,'' said $P11_{IC}$. $P29_{EC}$ justified this transfer by appeal to the division of labour. Humans are occupied with the task itself, so inference is properly the robot's contribution. They also recognised that what felt natural to produce was ``more complicated'' to interpret. Naturalness for the speaker and difficulty for the listener were understood as two descriptions of one utterance.
$P10_{EI}$ and $P23_{EC}$ described the one in a team that constantly gave indirect requests as a ``lazy communicator,'' and suggested that the Collaborator had to compensate for this laziness by devoting more effort to the team. The Collaborator's effort was also perceived in its explicit confirmations, which participants interpreted as a sign of cooperation.

However, the Collaborator's extra effort in the $Explicit$ condition was less preferred in some situations. In addition to the $Explicit \times Correct$ feedback reported above, $P16_{EI}$ reported that their sense of effort increased when observing the Collaborator's failures to understand in the low-severity task. The Collaborator's confirmations were seen as odd when participants believed their intent was quite obvious. They were generally regarded as unnecessary in the low-severity task, except in cases where the request was too vague and could yield multiple possible interpretations ($P1_{IC}$, $P7_{II}$). Moreover, the cost of time for confirming also made participants feel frustrated ($P2_{EI}$, $P25_{EI}$). Across these comments, the Collaborator was described as not life-like, but not merely machine-like either, and credited with a form of reasoning that was its own even when wrong.

\subsubsection{\textbf{Balanced Communication Effort}}
Participants appreciated the Collaborator's effort in confirming intents in the high-severity task under the $Explicit \times Incorrect$ condition ($P12_{EI}$, $P16_{EI}$, $P31_{EI}$). They believed the Collaborator was giving sufficient effort in interpreting and confirming, although incorrect in the first round. Confirmation also worked in the other direction and clarified the task for participants themselves. By restating vague requests as explicit ones, the Collaborator returned participants a specified version of their own intent ($P8_{EC}$, $P26_{EC}$).
``It's (Tiago) very predictable in a sense,'' said $P31_{EI}$.

\begin{quote}
$P31_{EI}$: ``He's (Tiago) very analytical and very careful, which is good, I think. Because if you work in a hospital, you need that kind of precision. He checks in with you again to make sure he understands well enough. Might be a bit frustrating for some people, but I think that's good for a robot.''
\end{quote}

Similarly, the increase in both robots' and humans' efforts was preferred in the high-severity task under the $Explicit \times Correct$ condition. Participants reported that the Collaborator was expected to be engaged and active in the interaction and to contribute more effort in the high-severity task through explicit confirmation ($P1_{IC}$, $P7_{II}$, $P23_{EC}$). Part of the reason was that misunderstanding was less tolerable in the high-severity task. Moreover, participants felt that the Collaborator responded in a rigid manner, which led them to condense their speech. They felt that they would spend more effort on giving direct requests in the high-severity task, although the Collaborator was interpreting correctly ($P9_{IC}$, $P26_{EC}$).

\begin{quote}
$P1_{IC}$: ``(The medical task) I think Tiago could be more explicit, that would be better. Like if Tiago said, `Oh, do you want me to mix the test tube again?' Then ask for confirmation.''
\end{quote}

\subsection{Task Severity Frames the Meaning of Extra Effort in Collaboration}
Participants correctly discerned task severity, though some said it had little personal impact because they treated the study as a simulation. High severity was associated with nervousness and caution. Participants reported being deliberately careful with their word choice in order to bridge common ground with the Collaborator, especially when referring to the precise amount of powder. They considered three potential keywords that reflected different attributes: ``30 grams'' for a semantic feature, ``white box'' for the most obvious appearance feature, and ``the middle one'' for a locative feature. In practice, participants mostly relied on semantic or appearance features, as these were believed to be the clearest and most precise. However, $P20_{EC}$ used the locative feature to test the Collaborator's ability to understand, implying that while locative reference was seen as a natural description, it was also regarded as the most implicit cue for a robot. Severity was therefore absorbed as lexical caution, which did not change what participants asked for, but changed how much work they did to secure it.
Furthermore, the Collaborator was expected to contribute more effort in the high-severity task through explicit confirmation, but regarded as unnecessary in the low-severity task.

Severity also shaped how the Collaborator's uptake performance affected participants' communication approaches. In low-severity tasks, participants predominantly used indirect requests, which felt smoother and more natural, whereas in high-severity tasks they relied more on clear, direct instructions. $P2_{EI}$ reported not only trying to speak faster but also expecting the Collaborator to respond more quickly when asking confirming questions in the high-severity task. 
\section{Discussion}
In this section, we discuss the impact of robots' communication explicitness, contextual intent interpretation, and task severity on grounding and communication effort in HRC. 

\subsection{Subtleties of the Grounding Process}
The quantitative results showed that humans' perception of grounding was mainly affected by whether the robot was able to interpret the intent correctly. This finding aligns with previous work~\cite{chai_collaborative_2014}. However, Explicitness showed no reliable effect on perceived grounding. We treat this as an inconclusive result. The explicitness comparison rests entirely on a between-subject crossing. Accordingly, our data neither confirm nor contradict the qualitative findings of~\citet{zhang_can_2025}. What we can state is descriptive. Grounding ratings were concentrated at the end points of the scale (see~\autoref{fig:clmm_plot}(a)), with participants tending to award a full score when the robot understood and completed the action, and a low score when it did not.
Moreover, in another study where a robot physically accomplished tasks with a human partner, team performance was not affected by the type of communication when the robot's skill level was high~\cite{lee_effect_2023}. Although some studies have used linguistic or corpus measures to quantify common ground, for instance, the ratio of definite references to the total number of words~\cite{bangerter_lexical_2020, wilkes-gibbs_coordinating_1992}, and the number of descriptive words or the number of new content words produced~\cite{clark_referring_1986, liu_referential_2021}, these are not applicable for measuring the grounding of action-oriented intents. Consequently, the subtle experience of grounding may be better reflected in qualitative reports.

Where the quantitative model is silent on explicitness, participants' interviews are not, and they form the basis for what we claim about it. The qualitative findings showed that common ground was seen as stable when the robot interpreted intent correctly, and as natural when it communicated implicitly, particularly when it was able to make inferences based on physical and environmental affordance. Similar results have been identified in research on human intent recognition~\cite{zhang_human-object_2022, wan_handmethat_2022}. Our findings add a bi-directional collaborative view to this work. Participants were still able to establish fragile common ground by investing additional communication effort. This extra effort humans put into communication is expected to repair the common ground, which eventually will improve efficiency~\cite{rouse_common_2005}. When collaborating with robots, some participants found this extra effort investment acceptable, while others did not and eventually gave up.\textbf{ When participants no longer attempt to ground, they begin to treat the robot more as a tool, giving it direct imperative orders rather than engaging with it as a collaborator~\cite{zhang_can_2025}.} Our findings show the critical role of correctness in the grounding process and the additional value of implicit robot responses, highlighting individual differences in humans' tolerance for maintaining and repairing fragile common ground with robots.

\subsection{When to Minimise, When to Share the Effort}
Both the quantitative and qualitative results showed that when the robot's responses were implicit or correct, participants experienced reduced communication effort. In these cases, they did not perceive a corresponding increase in the robot's effort. This pattern is consistent with the principle of least collaborative effort in HHC~\cite{clark_referring_1986}, where overall effort is minimised rather than shifted between partners. The qualitative findings further indicated that participants perceived the robot as more human-like under these conditions. 

When responses were explicit or incorrect, participants reported an increase in their own effort, with no comparable increase in perceived robot effort. 
Some participants intentionally created confirmation opportunities themselves (e.g. quote from $P28_{II}$) to ``regulate'' robots' responses. This made the collaboration lean more towards humans accommodating the robots' behaviour~\cite{reeves_opening_2025}, thereby creating an extra burden on humans. Only when the robot was both explicit and incorrect did participants perceive an increase in the robot's effort. As this pattern depends on a between-subject interaction, we present it as a directional finding in need of further exploration. Read at face value, it suggests that humans notice the robot's effort mainly when it explicitly asks for their help in correcting its understanding. This echoes work on robot failures, where explicit apologies measurably improved trust~\cite{zhang_sorry_2023}. 

The qualitative findings further illuminated this result. Participants felt that the robot's confirmations were necessary when the interpretations were incorrect, complementing findings from~\cite{smith_exploring_2024}. They interpreted these confirmations as evidence of the robot's effort to make sense of ambiguous human instructions, thereby contributing to the grounding process. A recent study on voice interfaces reported similar preferences, showing that humans favour explicit confirmation when performing high-purposefulness tasks and when confidence is low~\cite{wang_exploring_2025}. However, in low-severity tasks, most participants found confirmations unnecessary and tiring, as they already had high confidence in the established common ground, which aligns with conclusions from~\cite{reimann_survey_2024}. These findings are consistent with a joint orientation towards minimising overall effort in the team.

This orientation did not hold across conditions. When the Collaborator was both explicit and correct in the high-severity task, participants preferred its confirmations and the additional effort they required, even though common ground had already been established. \textbf{In such cases, participants appeared to be balancing extra effort across both partners in order to secure mutual understanding, rather than minimising it.} It also reflects a risk-averse strategy. Moreover, participants expected robots to confirm more quickly in high-severity tasks, which aligns with and complements findings of~\cite{smith_exploring_2024}. These patterns suggest that the principle of least collaborative effort is context-dependent in human-robot collaboration. Our findings extend this line of work to physically situated collaboration with robots and identify the severity of consequences as a condition under which communication effort is redistributed.

\subsection{Designing Robots to Compensate for Possible Misunderstandings}
An important conclusion from our results is that a robot's misunderstanding has a strong impact on both perceived grounding and human effort. Although researchers are developing state-of-the-art solutions for intent understanding~\cite{hoffman_inferring_2024}, the complexity of the world and task context makes misunderstandings inevitable. It is therefore critically important to consider what robots should do to compensate for breakdowns in grounding and team communication effort, taking the context into account.

First, when the confidence in interpretation accuracy is low, it is necessary to ask for explicit human confirmation in high-severity tasks. Such confirmations may not strongly influence humans' perception of their own effort, and they are highly valuable for establishing and repairing common ground. 
Second, in low-severity tasks, occasional incorrect actions without confirmation are acceptable if the action can be interrupted by the human. However, it may lead to extreme frustration and resentment when robots repeatedly misinterpret intent without asking for confirmation~\cite{reimann_survey_2024}. 
\textbf{Third, by analysing how humans issue requests, robots can infer humans' beliefs about the status of common ground.} When humans give indirect, ambiguous requests, they may perceive common ground as stable or the task as low in severity, expecting to minimise communication effort. A shift from direct to indirect requests may signal that they believe common ground has been established, whereas a shift from indirect to direct requests may indicate that they believe common ground needs repair. In the latter case, humans expect robots to take initiative in confirming actions~\cite{lyu_signaling_2025}. In high-severity tasks, however, even when humans give indirect requests, they may still expect confirmation from the robot, either to ensure understanding or to help them clarify their own understanding of the task.
Moreover, the robot can signal that common ground has been recovered by mirroring humans' language in referencing objects while performing correct actions. This behaviour indicates the robot's uptake, demonstrating that it has successfully understood and accepted the human's utterance~\cite{austin_how_1975}, thereby encouraging mutual re-establishment of understanding. It helps to return to a stable state of common ground, where human partners can once again rely on indirect requests and brief explanations to reduce overall effort~\cite{kiesler_fostering_2005}.
Finally, when interpretation confidence is high, implicit communication is generally the best choice. Frequent confirmations of correct interpretations can damage humans' sense of collaboration, shifting the interaction toward less natural, command-like communication, unless the task is high in severity. \textbf{To secure grounding in high-severity tasks, the overall cost of communication effort becomes acceptable, making explicit confirmation preferable regardless of interpretation accuracy.}

These implications differ in the strength of their support. The primacy of correctness is established by the quantitative model. The reading of request directness emerged from the interviews. The recommendations concerning explicitness and task severity draw on directional patterns in the model, elaborated by the interview data, and would benefit from further exploration.

\subsection{Limitations and Future Work}
Within the constraints of a lab study, the task simulation did not fully reflect the range of risks encountered in real life, but focused only on situational risk. Future work should investigate the moderating effects of other types of risk (e.g. relational risk) in order to examine how the findings of this study may generalise more broadly~\cite{stuck_importance_2022}. Moreover, participants were primarily recruited from university communities, with the majority being college students. This sample composition may limit the generalisability of the findings to broader age ranges or to populations with different professional backgrounds, levels of domain expertise, or prior exposure to collaborative robotic systems. Relatedly, while our sample size was determined by an a priori power analysis and exceeded the estimated minimum requirement, the interaction effects and in particular Explicitness by Correctness interaction, which rests entirely on a between-subject crossing, would benefit from replication with larger samples to confirm their robustness.

The use of a commercial LLM to generate the Collaborator's behaviour also warrants reflection, and several design choices were made to contain its influence on the results. Because LLMs are inevitably embedded with linguistic and social biases~\cite{kumar_no_2025}, the Collaborator's behaviour was not left to the model's defaults but was tightly constrained through condition-specific prompts with explicit behavioural rules, few-shot examples, and a fixed output format, all of which we release in full. Nevertheless, the model's conversational style inevitably reflects the norms of its training data, so the implicit and explicit behaviours studied here are one instantiation of these constructs rather than a culturally neutral one. To limit the stochasticity inherent to LLMs, the temperature was set to 0, the three core indirect request interactions were fully scripted between the two robots, and compliance with the condition rules was monitored throughout the study as described in \autoref{sec:architecture}. As a result, variation among participants within the same condition was confined to surface wording in the free exchanges, while the manipulated behaviours and the standardised exposures remained constant. All interactions in this study were recorded as ROSBags, so agents' logs can be reviewed post hoc if needed~\cite{zhang_rosannotator_2025}. Future work could also examine how operators interact with the LLM system. Regarding reproducibility, we scripted the standardised robot-robot exchanges and released the complete prompts and materials in the Supplementary Material, so the methodology is reproducible. Although we report the fixed model snapshot used to maximise transparency, this snapshot will eventually be deprecated from the API, as is standard practice for OpenAI models, meaning exact replication with the same model will not be possible indefinitely. Moreover, LLMs are advancing rapidly, and we hope future work can revisit this paradigm using more state-of-the-art models and self-hosted open-weight models, or traditional dialogue managers, to test whether the findings reported here hold as the underlying technology evolves. The latter two approaches trade some of the naturalness afforded by a commercial LLM for exact controllability, and would complement the ecological validity offered by the LLM-based implementation used here.

Lastly, due to the short duration of the study and participants' unfamiliarity with the robot's capabilities, some participants reported that they would increasingly use more natural indirect requests if given the opportunity to work with the robot over a longer period, which aligns with findings from~\cite{zhang_can_2025}. Future research should therefore explore long-term studies to better understand how humans transition from direct to indirect requests, and to test whether the findings remain consistent when humans are consistently the ones providing indirect requests to robots.
\section{Conclusion}
In this study, we investigated the effects of a robot's communication explicitness and the correctness of its intent interpretation on grounding and communication effort in human-robot collaboration, and how task severity moderated these effects. Interpretation correctness was the key predictor associated with perceived grounding, whereas explicitness, correctness, and severity jointly shaped how communication effort was perceived and distributed.

A robot that communicated more did not necessarily relieve its partner. Participants reported greater effort of their own when the robot responded explicitly, and this cost was not accompanied by any gain in perceived grounding, which was associated with interpretation correctness alone. The interview accounts locate this pattern in the low-severity task, where confirmations of intents that participants regarded as obvious were described as unnecessary and time-consuming, prompting some to front-load precise instructions in order to preempt them. Under high severity, the additional effort associated with explicit responses was weakened, and participants welcomed increased contributions from both partners as a safeguard against error. Designers should therefore not assume that maximising a robot's communicative contribution correspondingly reduces the human's, and the appropriate level of explicitness depends on the cost of an error. These observations extend the principle of least collaborative effort~\cite{clark_referring_1986}. Participants tended to adjust their joint effort with the robot to what an error would cost, a pattern we term balanced communication effort. These findings offer guidance for the design of communication systems for collaborative robots, including compensatory strategies robots may adopt when misunderstanding is foreseeable.

\begin{acks}
We gratefully acknowledge the support provided by the Melbourne Research Scholarship (University of Melbourne) and the Australian Research Council (Grant No. FT250100459, CE260100108).
\end{acks}

\bibliographystyle{ACM-Reference-Format}
\bibliography{references, references-2}

@article{leichtmann_moral_2024,
    title = {Moral reasoning in a digital age: blaming artificial intelligence for incorrect high-risk decisions},
    volume = {43},
    issn = {1936-4733},
    shorttitle = {Moral reasoning in a digital age},
    url = {https://doi.org/10.1007/s12144-024-06658-2},
    doi = {10.1007/s12144-024-06658-2},
    language = {en},
    number = {41},
    urldate = {2026-07-22},
    journal = {Current Psychology},
    author = {Leichtmann, Benedikt and Hinterreiter, Andreas and Humer, Christina and Ventura, Alfio and Streit, Marc and Mara, Martina},
    month = nov,
    year = {2024},
    pages = {32412--32421},
}

@article{leichtmann_teaming_2025,
	title = {Teaming with a {Robot} in {Mixed} {Reality}: {Dynamics} of {Trust}, {Self}-{Efficacy}, and {Mental} {Models} {Affected} by {Information} {Richness}},
	volume = {41},
	issn = {1044-7318},
	shorttitle = {Teaming with a {Robot} in {Mixed} {Reality}},
	url = {https://doi.org/10.1080/10447318.2024.2331878},
	doi = {10.1080/10447318.2024.2331878},
	number = {5},
	urldate = {2026-07-22},
	journal = {International Journal of Human–Computer Interaction},
	publisher = {Taylor \& Francis},
	author = {Leichtmann, Benedikt and Meneweger, Thomas and Busch, Christine and Reiterer, Bernhard and Meyer, Kathrin and Rammer, Daniel and Haring, Roland and Mara, Martina},
	month = mar,
	year = {2025},
	pages = {3166--3183},
}

@inproceedings{caine_local_2016,
	address = {New York, NY, USA},
	series = {{CHI} '16},
	title = {Local {Standards} for {Sample} {Size} at {CHI}},
	isbn = {978-1-4503-3362-7},
	url = {https://dl.acm.org/doi/10.1145/2858036.2858498},
	doi = {10.1145/2858036.2858498},
	urldate = {2026-07-23},
	booktitle = {Proceedings of the 2016 {CHI} {Conference} on {Human} {Factors} in {Computing} {Systems}},
	publisher = {Association for Computing Machinery},
	author = {Caine, Kelly},
	month = may,
	year = {2016},
	pages = {981--992},
}

@inproceedings{bilal_design_2026,
	address = {New York, NY, USA},
	series = {{DIS} '26},
	title = {Design and {Evaluation} of {AR}-{Based} {Real}-{Time} {Feedback} {System} for {Kinesthetic} {Robot} {Teaching}},
	isbn = {979-8-4007-2563-0},
	url = {https://dl.acm.org/doi/10.1145/3800645.3812951},
	doi = {10.1145/3800645.3812951},
	urldate = {2026-07-23},
	booktitle = {Proceedings of the 2026 {Designing} {Interactive} {Systems} {Conference}},
	publisher = {Association for Computing Machinery},
	author = {Bilal, Muhammad and Ratnayake, Tharaka Sachintha and Chacon, D. Antony and Lipovetzky, Nir and Oetomo, Denny and Johal, Wafa},
	month = jun,
	year = {2026},
	pages = {332--356},
}

@inproceedings{tabatabaei_oops_2026,
	address = {New York, NY, USA},
	series = {{CHI} '26},
	title = {Oops, {I} {Did} {It} {Again} ({But} {I} {Know} {It}): {Robot} {Failure} {Consistency} and {Awareness} in {Human}-{Robot} {Collaboration}},
	isbn = {979-8-4007-2278-3},
	shorttitle = {Oops, {I} {Did} {It} {Again} ({But} {I} {Know} {It})},
	url = {https://dl.acm.org/doi/10.1145/3772318.3790336},
	doi = {10.1145/3772318.3790336},
	urldate = {2026-07-23},
	booktitle = {Proceedings of the 2026 {CHI} {Conference} on {Human} {Factors} in {Computing} {Systems}},
	publisher = {Association for Computing Machinery},
	author = {Tabatabaei, Ramtin and Kostakos, Vassilis and Johal, Wafa},
	month = apr,
	year = {2026},
	pages = {1--18},
}

@article{kumle_estimating_2021,
	title = {Estimating power in (generalized) linear mixed models: {An} open introduction and tutorial in {R}},
	volume = {53},
	issn = {1554-3528},
	shorttitle = {Estimating power in (generalized) linear mixed models},
	url = {https://doi.org/10.3758/s13428-021-01546-0},
	doi = {10.3758/s13428-021-01546-0},
	language = {en},
	number = {6},
	urldate = {2026-07-24},
	journal = {Behavior Research Methods},
	author = {Kumle, Levi and Võ, Melissa L.-H. and Draschkow, Dejan},
	month = dec,
	year = {2021},
	pages = {2528--2543},
}

@inproceedings{victor_syiem_better_2026,
	address = {New York, NY, USA},
	series = {{CHI} '26},
	title = {Better {Assumptions}, {Stronger} {Conclusions}: {The} {Case} for {Ordinal} {Regression} in {HCI}},
	isbn = {979-8-4007-2278-3},
	shorttitle = {Better {Assumptions}, {Stronger} {Conclusions}},
	url = {https://dl.acm.org/doi/10.1145/3772318.3790821},
	doi = {10.1145/3772318.3790821},
	urldate = {2026-07-23},
	booktitle = {Proceedings of the 2026 {CHI} {Conference} on {Human} {Factors} in {Computing} {Systems}},
	publisher = {Association for Computing Machinery},
	author = {Victor Syiem, Brandon and Velloso, Eduardo},
	month = apr,
	year = {2026},
	pages = {1--21},
}

@article{maxwell_using_2010,
	title = {Using {Numbers} in {Qualitative} {Research}},
	volume = {16},
	copyright = {https://journals.sagepub.com/page/policies/text-and-data-mining-license},
	issn = {1077-8004, 1552-7565},
	url = {https://journals.sagepub.com/doi/10.1177/1077800410364740},
	doi = {10.1177/1077800410364740},
	language = {en},
	number = {6},
	urldate = {2026-08-03},
	journal = {Qualitative Inquiry},
	author = {Maxwell, Joseph A.},
	month = jul,
	year = {2010},
	pages = {475--482},

}

@article{braun_one_2021,
	title = {One size fits all? {What} counts as quality practice in (reflexive) thematic analysis?},
	volume = {18},
	issn = {1478-0887},
	shorttitle = {One size fits all?},
	url = {https://doi.org/10.1080/14780887.2020.1769238},
	doi = {10.1080/14780887.2020.1769238},
	number = {3},
	urldate = {2026-08-03},
	journal = {Qualitative Research in Psychology},
	publisher = {Routledge},
	author = {Braun, Virginia and Clarke, Victoria},
	month = jul,
	year = {2021},
	pages = {328--352},
}

@misc{kumar_no_2025,
	title = {No {LLM} is {Free} {From} {Bias}: {A} {Comprehensive} {Study} of {Bias} {Evaluation} in {Large} {Language} {Models}},
	shorttitle = {No {LLM} is {Free} {From} {Bias}},
	url = {http://arxiv.org/abs/2503.11985},
	doi = {10.48550/arXiv.2503.11985},
	urldate = {2026-08-03},
	publisher = {arXiv},
	author = {Kumar, Charaka Vinayak and Urlana, Ashok and Kanumolu, Gopichand and Garlapati, Bala Mallikarjunarao and Mishra, Pruthwik},
	month = may,
	year = {2025},
	note = {arXiv:2503.11985 [cs.CL]}
}

@article{searle1980population,
  title={Population marginal means in the linear model: an alternative to least squares means},
  author={Searle, Shayle R and Speed, F Michael and Milliken, George A},
  journal={The American Statistician},
  volume={34},
  number={4},
  pages={216--221},
  year={1980},
  publisher={Taylor \& Francis}
}

@article{green_simr_2016,
	title = {{SIMR}: an {R} package for power analysis of generalized linear mixed models by simulation},
	volume = {7},
	copyright = {http://onlinelibrary.wiley.com/termsAndConditions\#vor},
	issn = {2041-210X, 2041-210X},
	doi = {10.1111/2041-210X.12504},
	language = {en},
	number = {4},
	urldate = {2026-10-16},
	journal = {Methods in Ecology and Evolution},
	author = {Green, Peter and MacLeod, Catriona J.},
	editor = {Nakagawa, Shinichi},
	month = apr,
	year = {2016},
	pages = {493--498}
}

@article{faul_statistical_2009,
	title = {Statistical power analyses using {G}*{Power} 3.1: {Tests} for correlation and regression analyses},
	volume = {41},
	issn = {1554-3528},
	shorttitle = {Statistical power analyses using {G}*{Power} 3.1},
	url = {https://doi.org/10.3758/BRM.41.4.1149},
	doi = {10.3758/BRM.41.4.1149},
	language = {en},
	number = {4},
	urldate = {2026-10-19},
	journal = {Behavior Research Methods},
	author = {Faul, Franz and Erdfelder, Edgar and Buchner, Axel and Lang, Albert-Georg},
	month = nov,
	year = {2009},
	pages = {1149--1160}
}

@article{obaigbena_ai_2024_2,
	title = {{AI} and human-robot interaction: {A} review of recent advances and challenges},
	volume = {18},
	issn = {2582-4597, 2582-4597},
	shorttitle = {{AI} and human-robot interaction},
	url = {https://gsconlinepress.com/journals/gscarr/content/ai-and-human-robot-interaction-review-recent-advances-and-challenges},
	doi = {10.30574/gscarr.2024.18.2.0070},
	number = {2},
	urldate = {2025-09-02},
	journal = {GSC Advanced Research and Reviews},
	author = {Obaigbena, Alexander and Lottu, Oluwaseun Augustine and Ugwuanyi, Ejike David and Jacks, Boma Sonimitiem and Sodiya, Enoch Oluwademilade and Daraojimba, Obinna Donald},
	year = {2024},
	pages = {321--330}
}

@article{garrido-jurado_automatic_2014,
	title = {Automatic generation and detection of highly reliable fiducial markers under occlusion},
	volume = {47},
	issn = {0031-3203},
	url = {https://www.sciencedirect.com/science/article/pii/S0031320314000235},
	doi = {10.1016/j.patcog.2014.01.005},
	number = {6},
	urldate = {2026-10-19},
	journal = {Pattern Recognition},
	author = {Garrido-Jurado, S. and Muñoz-Salinas, R. and Madrid-Cuevas, F. J. and Marín-Jiménez, M. J.},
	month = jun,
	year = {2014},
	pages = {2280--2292}
}

@inproceedings{zhang_rosannotator_2025,
	title = {{ROSAnnotator}: {A} {Web} {Application} for {ROSBag} {Data} {Analysis} in {Human}-{Robot} {Interaction}},
	shorttitle = {{ROSAnnotator}},
	url = {https://ieeexplore.ieee.org/abstract/document/10974254},
	doi = {10.1109/HRI61500.2025.10974254},
	booktitle = {2025 20th {ACM}/{IEEE} {International} {Conference} on {Human}-{Robot} {Interaction} ({HRI})},
	author = {Zhang, Yan and Li, Haoqi and Tabatabaei, Ramtin and Johal, Wafa},
	month = mar,
	year = {2025},
	pages = {1099--1103}
}

@misc{desmond_exploring_2024,
	title = {Exploring {Prompt} {Engineering} {Practices} in the {Enterprise}},
	url = {http://arxiv.org/abs/2403.08950},
	doi = {10.48550/arXiv.2403.08950},
	urldate = {2025-09-03},
	publisher = {arXiv},
	author = {Desmond, Michael and Brachman, Michelle},
	month = mar,
	year = {2024},
}

@inproceedings{kontogiorgos_behavioural_2020,
	address = {New York, NY, USA},
	series = {{HRI} '20},
	title = {Behavioural {Responses} to {Robot} {Conversational} {Failures}},
	isbn = {978-1-4503-6746-2},
	url = {https://dl.acm.org/doi/10.1145/3319502.3374782},
	doi = {10.1145/3319502.3374782},
	urldate = {2025-09-02},
	booktitle = {Proceedings of the 2020 {ACM}/{IEEE} {International} {Conference} on {Human}-{Robot} {Interaction}},
	publisher = {Association for Computing Machinery},
	author = {Kontogiorgos, Dimosthenis and Pereira, Andre and Sahindal, Boran and van Waveren, Sanne and Gustafson, Joakim},
	month = mar,
	year = {2020},
	pages = {53--62},
}

@article{xie_risk_2003,
	title = {Risk perception and risky choice: {Situational}, informational and dispositional effects},
	volume = {6},
	issn = {1467-839X},
	shorttitle = {Risk perception and risky choice},
	url = {https://onlinelibrary.wiley.com/doi/abs/10.1111/1467-839X.t01-1-00015},
	doi = {10.1111/1467-839X.t01-1-00015},
	language = {en},
	number = {2},
	urldate = {2025-09-02},
	journal = {Asian Journal of Social Psychology},
	author = {Xie, Xiao-Fei and Wang, X. T. (XiaoTian)},
	year = {2003},
	pages = {117--132},
}

@inproceedings{torrey_effects_2006,
	address = {Salt Lake City Utah USA},
	title = {Effects of adaptive robot dialogue on information exchange and social relations},
	isbn = {978-1-59593-294-5},
	url = {https://dl.acm.org/doi/10.1145/1121241.1121264},
	doi = {10.1145/1121241.1121264},
	language = {en},
	urldate = {2025-09-29},
	booktitle = {Proceedings of the 1st {ACM} {SIGCHI}/{SIGART} conference on {Human}-robot interaction},
	publisher = {ACM},
	author = {Torrey, Cristen and Powers, Aaron and Marge, Matthew and Fussell, Susan R. and Kiesler, Sara},
	month = mar,
	year = {2006},
	pages = {126--133},
}

@article{hoffman_inferring_2024,
	title = {Inferring {Human} {Intent} and {Predicting} {Human} {Action} in {Human}–{Robot} {Collaboration}},
	volume = {7},
	issn = {2573-5144},
	url = {https://www.annualreviews.org/content/journals/10.1146/annurev-control-071223-105834},
	doi = {10.1146/annurev-control-071223-105834},
	language = {zh-CN},
	number = {Volume 7, 2024},
	urldate = {2025-04-01},
	journal = {Annual Review of Control, Robotics, and Autonomous Systems},
	author = {Hoffman, Guy and Bhattacharjee, Tapomayukh and Nikolaidis, Stefanos},
	month = jul,
	year = {2024},
	pages = {73--95},
}

@article{zheng_putting_2024,
	title = {Putting it {Together}, {Together}},
	volume = {48},
	copyright = {© 2024 Cognitive Science Society LLC.},
	issn = {1551-6709},
	url = {https://onlinelibrary.wiley.com/doi/abs/10.1111/cogs.13405},
	doi = {10.1111/cogs.13405},
	language = {en},
	number = {2},
	urldate = {2025-09-10},
	journal = {Cognitive Science},
	author = {Zheng, Chen and Tversky, Barbara},
	year = {2024},
	pages = {e13405},
}

@book{austin_how_1975,
	title = {How to {Do} {Things} with {Words}: {Second} {Edition}},
	isbn = {978-0-674-25211-0},
	shorttitle = {How to {Do} {Things} with {Words}},
	language = {en},
	publisher = {Harvard University Press},
	author = {Austin, J. L.},
	month = apr,
	year = {1975},
}

@inproceedings{gao_physically_2024,
	title = {Physically {Grounded} {Vision}-{Language} {Models} for {Robotic} {Manipulation}},
	url = {https://ieeexplore.ieee.org/document/10610090/},
	doi = {10.1109/ICRA57147.2024.10610090},
	language = {en-US},
	urldate = {2025-09-09},
	booktitle = {2024 {IEEE} {International} {Conference} on {Robotics} and {Automation} ({ICRA})},
	author = {Gao, Jensen and Sarkar, Bidipta and Xia, Fei and Xiao, Ted and Wu, Jiajun and Ichter, Brian and Majumdar, Anirudha and Sadigh, Dorsa},
	month = may,
	year = {2024},
	pages = {12462--12469},
}

@inproceedings{wanna_lets_2025,
	title = {Let's {Talk} {About} {Language}! {Investigating} {Linguistic} {Diversity} in {Embodied} {AI} {Datasets}},
	url = {https://openreview.net/forum?id=wWWKPNz6GJ},
	language = {en},
	urldate = {2025-09-09},
	author = {Wanna, Selma Liliane and Luhtaru, Agnes and Barron, Ryan and Salfity, Jonathan and Moore, Juston and Matuszek, Cynthia and Pryor, Mitch},
	month = may,
	year = {2025},
}

@misc{jiang_llms_2024,
	title = {{LLMs} for {Robotic} {Object} {Disambiguation}},
	url = {http://arxiv.org/abs/2401.03388},
	doi = {10.48550/arXiv.2401.03388},
	language = {en-US},
	urldate = {2025-09-09},
	publisher = {arXiv},
	author = {Jiang, Connie and Xu, Yiqing and Hsu, David},
	month = jan,
	year = {2024},
}

@article{chai_collaborative_2016,
	title = {Collaborative {Language} {Grounding} {Toward} {Situated} {Human}‐{Robot} {Dialogue}},
	volume = {37},
	copyright = {http://onlinelibrary.wiley.com/termsAndConditions\#am},
	issn = {0738-4602, 2371-9621},
	url = {https://onlinelibrary.wiley.com/doi/10.1609/aimag.v37i4.2684},
	doi = {10.1609/aimag.v37i4.2684},
	language = {en},
	number = {4},
	urldate = {2024-05-16},
	journal = {AI Magazine},
	author = {Chai, Joyce Y. and Fang, Rui and Liu, Changsong and She, Lanbo},
	month = dec,
	year = {2016},
	pages = {32--45},
}

@inproceedings{allgeuer_when_2024,
	address = {Cham},
	title = {When {Robots} {Get} {Chatty}: {Grounding} {Multimodal} {Human}-{Robot} {Conversation} and {Collaboration}},
	isbn = {978-3-031-72341-4},
	shorttitle = {When {Robots} {Get} {Chatty}},
	doi = {10.1007/978-3-031-72341-4_21},
	language = {en},
	booktitle = {Artificial {Neural} {Networks} and {Machine} {Learning} – {ICANN} 2024},
	publisher = {Springer Nature Switzerland},
	author = {Allgeuer, Philipp and Ali, Hassan and Wermter, Stefan},
	editor = {Wand, Michael and Malinovská, Kristína and Schmidhuber, Jürgen and Tetko, Igor V.},
	year = {2024},
	pages = {306--321},
}

@inproceedings{chai_collaborative_2014,
	address = {Bielefeld Germany},
	title = {Collaborative effort towards common ground in situated human-robot dialogue},
	isbn = {978-1-4503-2658-2},
	url = {https://dl.acm.org/doi/10.1145/2559636.2559677},
	doi = {10.1145/2559636.2559677},
	language = {en},
	urldate = {2025-03-11},
	booktitle = {Proceedings of the 2014 {ACM}/{IEEE} international conference on {Human}-robot interaction},
	publisher = {ACM},
	author = {Chai, Joyce Y. and She, Lanbo and Fang, Rui and Ottarson, Spencer and Littley, Cody and Liu, Changsong and Hanson, Kenneth},
	month = mar,
	year = {2014},
	pages = {33--40},
}

@inproceedings{liang_implicit_2019,
	address = {Glasgow Scotland Uk},
	title = {Implicit {Communication} of {Actionable} {Information} in {Human}-{AI} teams},
	isbn = {978-1-4503-5970-2},
	url = {https://dl.acm.org/doi/10.1145/3290605.3300325},
	doi = {10.1145/3290605.3300325},
	language = {en},
	urldate = {2025-03-11},
	booktitle = {Proceedings of the 2019 {CHI} {Conference} on {Human} {Factors} in {Computing} {Systems}},
	publisher = {ACM},
	author = {Liang, Claire and Proft, Julia and Andersen, Erik and Knepper, Ross A.},
	month = may,
	year = {2019},
	pages = {1--13},
}

@inproceedings{knepper_implicit_2017,
	address = {Vienna Austria},
	title = {Implicit {Communication} in a {Joint} {Action}},
	isbn = {978-1-4503-4336-7},
	url = {https://dl.acm.org/doi/10.1145/2909824.3020226},
	doi = {10.1145/2909824.3020226},
	language = {en},
	urldate = {2025-03-11},
	booktitle = {Proceedings of the 2017 {ACM}/{IEEE} {International} {Conference} on {Human}-{Robot} {Interaction}},
	publisher = {ACM},
	author = {Knepper, Ross A. and Mavrogiannis, Christoforos I. and Proft, Julia and Liang, Claire},
	month = mar,
	year = {2017},
	pages = {283--292},
}

@inproceedings{kontogiorgos_towards_2020,
	address = {Cambridge United Kingdom},
	title = {Towards {Adaptive} and {Least}-{Collaborative}-{Effort} {Social} {Robots}},
	isbn = {978-1-4503-7057-8},
	url = {https://dl.acm.org/doi/10.1145/3371382.3378249},
	doi = {10.1145/3371382.3378249},
	language = {en},
	urldate = {2025-09-07},
	booktitle = {Companion of the 2020 {ACM}/{IEEE} {International} {Conference} on {Human}-{Robot} {Interaction}},
	publisher = {ACM},
	author = {Kontogiorgos, Dimosthenis and Pelikan, Hannah R. M.},
	month = mar,
	year = {2020},
	pages = {311--313},
}

@article{belhassein_addressing_2022,
	title = {Addressing joint action challenges in {HRI}: {Insights} from psychology and philosophy},
	volume = {222},
	shorttitle = {Addressing joint action challenges in {HRI}},
	doi = {10.1016/j.actpsy.2021.103476},
	language = {en-US},
	journal = {Acta Psychologica},
	author = {Belhassein, Kathleen and Fernández Castro, Víctor and Mayima, Amandine and Aurélie, Clodic and Pacherie, Elisabeth and Michèle, Guidetti and Alami, Rachid and Hélène, Cochet},
	month = feb,
	year = {2022},
	pages = {103476},
}

@article{reimann_survey_2024,
	title = {A {Survey} on {Dialogue} {Management} in {Human}-robot {Interaction}},
	volume = {13},
	issn = {2573-9522},
	url = {https://dl.acm.org/doi/10.1145/3648605},
	doi = {10.1145/3648605},
	language = {en},
	number = {2},
	urldate = {2025-09-07},
	journal = {ACM Transactions on Human-Robot Interaction},
	author = {Reimann, Merle M. and Kunneman, Florian A. and Oertel, Catharine and Hindriks, Koen V.},
	month = jun,
	year = {2024},
	pages = {1--22},
}

@misc{yu_mixed-initiative_2025,
	title = {Mixed-{Initiative} {Dialog} for {Human}-{Robot} {Collaborative} {Manipulation}},
	url = {http://arxiv.org/abs/2508.05535},
	doi = {10.48550/arXiv.2508.05535},
	urldate = {2025-09-07},
	publisher = {arXiv},
	author = {Yu, Albert and Li, Chengshu and Macesanu, Luca and Balaji, Arnav and Ray, Ruchira and Mooney, Raymond and Martín-Martín, Roberto},
	month = aug,
	year = {2025},
}

@inproceedings{brandstetter_persistent_2017,
	address = {Vienna Austria},
	title = {Persistent {Lexical} {Entrainment} in {HRI}},
	isbn = {978-1-4503-4336-7},
	url = {https://dl.acm.org/doi/10.1145/2909824.3020257},
	doi = {10.1145/2909824.3020257},
	language = {en},
	urldate = {2025-09-07},
	booktitle = {Proceedings of the 2017 {ACM}/{IEEE} {International} {Conference} on {Human}-{Robot} {Interaction}},
	publisher = {ACM},
	author = {Brandstetter, Jürgen and Beckner, Clay and Sandoval, Eduardo Benitez and Bartneck, Christoph},
	month = mar,
	year = {2017},
	pages = {63--72},
}

@inproceedings{iio_lexical_2009,
	title = {Lexical entrainment in human-robot interaction: {Can} robots entrain human vocabulary?},
	issn = {2153-0866},
	shorttitle = {Lexical entrainment in human-robot interaction},
	url = {https://ieeexplore.ieee.org/abstract/document/5354149},
	doi = {10.1109/IROS.2009.5354149},
	urldate = {2025-09-07},
	booktitle = {2009 {IEEE}/{RSJ} {International} {Conference} on {Intelligent} {Robots} and {Systems}},
	author = {Iio, Takamasa and Shiomi, Masahiro and Shinozawa, Kazuhiko and Miyashita, Takahiro and Akimoto, Takaaki and Hagita, Norihiro},
	month = oct,
	year = {2009},
	pages = {3727--3734},
}

@article{iio_lexical_2015,
	title = {Lexical {Entrainment} in {Human} {Robot} {Interaction}},
	volume = {7},
	issn = {1875-4805},
	url = {https://doi.org/10.1007/s12369-014-0255-x},
	doi = {10.1007/s12369-014-0255-x},
	language = {en},
	number = {2},
	urldate = {2025-09-07},
	journal = {International Journal of Social Robotics},
	author = {Iio, Takamasa and Shiomi, Masahiro and Shinozawa, Kazuhiko and Shimohara, Katsunori and Miki, Mitsunori and Hagita, Norihiro},
	month = apr,
	year = {2015},
	pages = {253--263},
}

@article{marge_miscommunication_2019,
	title = {Miscommunication {Detection} and {Recovery} in {Situated} {Human}–{Robot} {Dialogue}},
	volume = {9},
	issn = {2160-6455, 2160-6463},
	url = {https://dl.acm.org/doi/10.1145/3237189},
	doi = {10.1145/3237189},
	language = {en},
	number = {1},
	urldate = {2025-09-07},
	journal = {ACM Transactions on Interactive Intelligent Systems},
	author = {Marge, Matthew and Rudnicky, Alexander I.},
	month = mar,
	year = {2019},
	pages = {1--40},
}

@inproceedings{devin_implemented_2016,
	title = {An implemented theory of mind to improve human-robot shared plans execution},
	issn = {2167-2148},
	url = {https://ieeexplore.ieee.org/abstract/document/7451768},
	doi = {10.1109/HRI.2016.7451768},
	urldate = {2025-09-07},
	booktitle = {2016 11th {ACM}/{IEEE} {International} {Conference} on {Human}-{Robot} {Interaction} ({HRI})},
	author = {Devin, Sandra and Alami, Rachid},
	month = mar,
	year = {2016},
	pages = {319--326},
}

@article{glenberg_grounding_2002,
	title = {Grounding language in action},
	volume = {9},
	issn = {1531-5320},
	url = {https://doi.org/10.3758/BF03196313},
	doi = {10.3758/BF03196313},
	language = {en},
	number = {3},
	urldate = {2025-08-31},
	journal = {Psychonomic Bulletin \& Review},
	author = {Glenberg, Arthur M. and Kaschak, Michael P.},
	month = sep,
	year = {2002},
	pages = {558--565},
}

@article{pickering_toward_2004,
	title = {Toward a mechanistic psychology of dialogue},
	volume = {27},
	issn = {0140-525X, 1469-1825},
	url = {https://www.cambridge.org/core/journals/behavioral-and-brain-sciences/article/abs/toward-a-mechanistic-psychology-of-dialogue/83442BA495E0D5F81BDB615E4109DBD2},
	doi = {10.1017/S0140525X04000056},
	language = {en},
	number = {2},
	urldate = {2025-08-31},
	journal = {Behavioral and Brain Sciences},
	author = {Pickering, Martin J. and Garrod, Simon},
	month = apr,
	year = {2004},
	pages = {169--190},
}

@article{visser_model_2014,
	title = {A model for incremental grounding in spoken dialogue systems},
	volume = {8},
	issn = {1783-8738},
	url = {https://doi.org/10.1007/s12193-013-0147-7},
	doi = {10.1007/s12193-013-0147-7},
	language = {en},
	number = {1},
	urldate = {2025-08-31},
	journal = {Journal on Multimodal User Interfaces},
	author = {Visser, Thomas and Traum, David and DeVault, David and op den Akker, Rieks},
	month = mar,
	year = {2014},
	pages = {61--73},
}

@phdthesis{traum_computational_1995,
	address = {United States -- New York},
	type = {Ph.{D}.},
	title = {A computational theory of grounding in natural language conversation},
	copyright = {Database copyright ProQuest LLC; ProQuest does not claim copyright in the individual underlying works.},
	isbn = {979-8-6417-7097-0},
	url = {https://www.proquest.com/docview/304219107/abstract/E33660ECB28345E7PQ/1},
	language = {English},
	urldate = {2025-08-31},
	school = {University of Rochester},
	author = {Traum, David Rood},
	year = {1995},
}

@book{clark_using_1996,
	title = {Using {Language}},
	isbn = {978-0-521-56745-9},
	language = {en},
	publisher = {Cambridge University Press},
	author = {Clark, Herbert H.},
	month = may,
	year = {1996},
}

@article{clark_contributing_1989,
	title = {Contributing to discourse},
	volume = {13},
	issn = {0364-0213},
	url = {https://www.sciencedirect.com/science/article/pii/0364021389900086},
	doi = {10.1016/0364-0213(89)90008-6},
	language = {en-US},
	number = {2},
	urldate = {2025-08-31},
	journal = {Cognitive Science},
	author = {Clark, Herbert H. and Schaefer, Edward F.},
	month = apr,
	year = {1989},
	pages = {259--294},
}

@incollection{clark_grounding_1991,
	address = {Washington, DC, US},
	title = {Grounding in communication},
	isbn = {978-1-55798-121-9},
	doi = {10.1037/10096-006},
	language = {en-US},
	booktitle = {Perspectives on socially shared cognition},
	publisher = {American Psychological Association},
	author = {Clark, Herbert H. and Brennan, Susan E.},
	year = {1991},
	pages = {127--149},
}

@article{clark_referring_1986,
	title = {Referring as a collaborative process},
	volume = {22},
	issn = {0010-0277},
	url = {https://www.sciencedirect.com/science/article/pii/0010027786900107},
	doi = {10.1016/0010-0277(86)90010-7},
	language = {en-US},
	number = {1},
	urldate = {2025-08-31},
	journal = {Cognition},
	author = {Clark, Herbert H. and Wilkes-Gibbs, Deanna},
	month = feb,
	year = {1986},
	pages = {1--39},
}

@article{hancock_meta-analysis_2011,
	title = {A {Meta}-{Analysis} of {Factors} {Affecting} {Trust} in {Human}-{Robot} {Interaction}},
	volume = {53},
	issn = {0018-7208},
	url = {https://doi.org/10.1177/0018720811417254},
	doi = {10.1177/0018720811417254},
	language = {EN},
	number = {5},
	urldate = {2025-09-02},
	journal = {Human Factors},
	author = {Hancock, Peter A. and Billings, Deborah R. and Schaefer, Kristin E. and Chen, Jessie Y. C. and de Visser, Ewart J. and Parasuraman, Raja},
	month = oct,
	year = {2011},
	pages = {517--527},
}

@incollection{bhat_value_2024,
	address = {Cham},
	title = {Value {Alignment} and {Trust} in {Human}-{Robot} {Interaction}: {Insights} from {Simulation} and {User} {Study}},
	isbn = {978-3-031-66656-8},
	shorttitle = {Value {Alignment} and {Trust} in {Human}-{Robot} {Interaction}},
	url = {https://doi.org/10.1007/978-3-031-66656-8_3},
	doi = {10.1007/978-3-031-66656-8_3},
	language = {en},
	urldate = {2025-09-02},
	booktitle = {Discovering the {Frontiers} of {Human}-{Robot} {Interaction}: {Insights} and {Innovations} in {Collaboration}, {Communication}, and {Control}},
	publisher = {Springer Nature Switzerland},
	author = {Bhat, Shreyas and Lyons, Joseph B. and Shi, Cong and Yang, X. Jessie},
	editor = {Vinjamuri, Ramana},
	year = {2024},
	pages = {39--63},
}

@article{alhaji_trust_2021,
	title = {Trust {Dynamics} and {Verbal} {Assurances} in {Human} {Robot} {Physical} {Collaboration}},
	volume = {4},
	issn = {2624-8212},
	url = {https://www.frontiersin.org/journals/artificial-intelligence/articles/10.3389/frai.2021.703504/full},
	doi = {10.3389/frai.2021.703504},
	language = {en-US},
	urldate = {2025-09-02},
	journal = {Frontiers in Artificial Intelligence},
	author = {Alhaji, Basel and Prilla, Michael and Rausch, Andreas},
	month = jul,
	year = {2021},
}

@inproceedings{schombs_robot-assisted_2024,
	address = {New York, NY, USA},
	series = {{CHI} '24},
	title = {Robot-{Assisted} {Decision}-{Making}: {Unveiling} the {Role} of {Uncertainty} {Visualisation} and {Embodiment}},
	isbn = {979-8-4007-0330-0},
	shorttitle = {Robot-{Assisted} {Decision}-{Making}},
	url = {https://dl.acm.org/doi/10.1145/3613904.3642911},
	doi = {10.1145/3613904.3642911},
	urldate = {2025-09-09},
	booktitle = {Proceedings of the 2024 {CHI} {Conference} on {Human} {Factors} in {Computing} {Systems}},
	publisher = {Association for Computing Machinery},
	author = {Schömbs, Sarah and Pareek, Saumya and Goncalves, Jorge and Johal, Wafa},
	month = may,
	year = {2024},
	pages = {1--16},
}

@article{grasse_speech_2021,
	title = {Speech {Interaction} to {Control} a {Hands}-{Free} {Delivery} {Robot} for {High}-{Risk} {Health} {Care} {Scenarios}},
	volume = {8},
	issn = {2296-9144},
	url = {https://www.frontiersin.org/journals/robotics-and-ai/articles/10.3389/frobt.2021.612750/full},
	doi = {10.3389/frobt.2021.612750},
	language = {en-US},
	urldate = {2025-09-10},
	journal = {Frontiers in Robotics and AI},
	author = {Grasse, Lukas and Boutros, Sylvain J. and Tata, Matthew S.},
	month = apr,
	year = {2021},
}

@inproceedings{smith_exploring_2024,
	title = {Exploring the {Impact} of {Confirmation} and {Interaction} {During} {Human}-{Robot} {Collaboration} with a {Proactive} {Robot} {Assistant}},
	issn = {1944-9437},
	url = {https://ieeexplore.ieee.org/abstract/document/10731366},
	doi = {10.1109/RO-MAN60168.2024.10731366},
	language = {en-US},
	urldate = {2025-09-02},
	booktitle = {2024 33rd {IEEE} {International} {Conference} on {Robot} and {Human} {Interactive} {Communication} ({ROMAN})},
	author = {Smith, Ronnie and Dragone, Mauro},
	month = aug,
	year = {2024},
	pages = {2007--2014},
}

@article{stuck_importance_2022,
	title = {The importance of incorporating risk into human-automation trust},
	volume = {23},
	issn = {1463-922X},
	url = {https://doi.org/10.1080/1463922X.2021.1975170},
	doi = {10.1080/1463922X.2021.1975170},
	language = {en-US},
	number = {4},
	urldate = {2025-09-09},
	journal = {Theoretical Issues in Ergonomics Science},
	author = {Stuck, Rachel E. and Tomlinson, Brianna J. and Walker, Bruce N.},
	month = jul,
	year = {2022},
	pages = {500--516},
}

@article{wang_exploring_2025,
	title = {Exploring {Confirmation} {Strategies} for {Voice} {Interaction} in {Multi}-{Tasking} {Scenario}},
	issn = {1875-4805},
	url = {https://doi.org/10.1007/s12369-025-01302-w},
	doi = {10.1007/s12369-025-01302-w},
	language = {en},
	urldate = {2025-09-08},
	journal = {International Journal of Social Robotics},
	author = {Wang, Junfeng and Xu, Zhiyu and Zhai, Weimin and Xu, Fei},
	month = aug,
	year = {2025},
}

@article{zhang_sorry_2023,
	title = {“{Sorry}, it was my fault”: {Repairing} trust in human-robot interactions},
	volume = {175},
	issn = {1071-5819},
	shorttitle = {“{Sorry}, it was my fault”},
	url = {https://www.sciencedirect.com/science/article/pii/S107158192300037X},
	doi = {10.1016/j.ijhcs.2023.103031},
	urldate = {2025-09-08},
	journal = {International Journal of Human-Computer Studies},
	author = {Zhang, Xinyi and Lee, Sun Kyong and Kim, Whani and Hahn, Sowon},
	month = jul,
	year = {2023},
	pages = {103031},
}

@incollection{rouse_common_2005,
	edition = {1},
	title = {Common {Ground} and {Coordination} in {Joint} {Activity}},
	isbn = {978-0-471-68163-2 978-0-471-73944-9},
	url = {https://onlinelibrary.wiley.com/doi/10.1002/0471739448.ch6},
	doi = {10.1002/0471739448.ch6},
	language = {en},
	urldate = {2025-09-08},
	booktitle = {Organizational {Simulation}},
	publisher = {Wiley},
	author = {Klein, Gary and Feltovich, Paul J. and Bradshaw, Jeffrey M. and Woods, David D.},
	editor = {Rouse, William B. and Boff, Kenneth R.},
	month = may,
	year = {2005},
	pages = {139--184},
}

@article{wan_handmethat_2022,
	title = {{HandMeThat}: {Human}-{Robot} {Communication} in {Physical} and {Social} {Environments}},
	volume = {35},
	shorttitle = {{HandMeThat}},
	url = {https://proceedings.neurips.cc/paper_files/paper/2022/hash/4eb33c53ed5b14ce9028309431f565cc-Abstract-Datasets_and_Benchmarks.html},
	language = {en},
	urldate = {2025-09-08},
	journal = {Advances in Neural Information Processing Systems},
	author = {Wan, Yanming and Mao, Jiayuan and Tenenbaum, Josh},
	month = dec,
	year = {2022},
	pages = {12014--12026},
}

@article{zhang_human-object_2022,
	title = {Human-object integrated assembly intention recognition for context-aware human-robot collaborative assembly},
	volume = {54},
	issn = {1474-0346},
	url = {https://www.sciencedirect.com/science/article/pii/S1474034622002506},
	doi = {10.1016/j.aei.2022.101792},
	urldate = {2025-09-08},
	journal = {Advanced Engineering Informatics},
	author = {Zhang, Yaqian and Ding, Kai and Hui, Jizhuang and Lv, Jingxiang and Zhou, Xueliang and Zheng, Pai},
	month = oct,
	year = {2022},
	pages = {101792},
}

@inproceedings{lee_effect_2023,
	address = {Stockholm Sweden},
	title = {The {Effect} of {Robot} {Skill} {Level} and {Communication} in {Rapid}, {Proximate} {Human}-{Robot} {Collaboration}},
	isbn = {978-1-4503-9964-7},
	url = {https://dl.acm.org/doi/10.1145/3568162.3577002},
	doi = {10.1145/3568162.3577002},
	language = {en},
	urldate = {2025-09-08},
	booktitle = {Proceedings of the 2023 {ACM}/{IEEE} {International} {Conference} on {Human}-{Robot} {Interaction}},
	publisher = {ACM},
	author = {Lee, Kin Man and Krishna, Arjun and Zaidi, Zulfiqar and Paleja, Rohan and Chen, Letian and Hedlund-Botti, Erin and Schrum, Mariah and Gombolay, Matthew},
	month = mar,
	year = {2023},
	pages = {261--270},
}

@article{liu_referential_2021,
	title = {Referential {Communication} {Between} {Friends} and {Strangers} in the {Wild}},
	volume = {12},
	copyright = {Copyright (c)},
	issn = {2152-9620},
	url = {https://journals.uic.edu/ojs/index.php/dad/article/view/11397},
	doi = {10.5210/dad.2021.103},
	language = {en},
	number = {1},
	urldate = {2025-09-08},
	journal = {Dialogue \& Discourse},
	author = {Liu, Kris and D'Arcey, J. Trevor and Walker, Marilyn and Tree, Jean Fox},
	month = apr,
	year = {2021},
	pages = {45--72},
}

@article{wilkes-gibbs_coordinating_1992,
	title = {Coordinating beliefs in conversation},
	volume = {31},
	issn = {0749-596X},
	url = {https://www.sciencedirect.com/science/article/pii/0749596X9290010U},
	doi = {10.1016/0749-596X(92)90010-U},
	number = {2},
	urldate = {2025-09-08},
	journal = {Journal of Memory and Language},
	author = {Wilkes-Gibbs, Deanna and Clark, Herbert H},
	month = apr,
	year = {1992},
	pages = {183--194},
}

@article{bangerter_lexical_2020,
	title = {Lexical entrainment without conceptual pacts? {Revisiting} the matching task},
	volume = {114},
	issn = {0749-596X},
	shorttitle = {Lexical entrainment without conceptual pacts?},
	url = {https://www.sciencedirect.com/science/article/pii/S0749596X20300437},
	doi = {10.1016/j.jml.2020.104129},
	urldate = {2025-09-08},
	journal = {Journal of Memory and Language},
	author = {Bangerter, Adrian and Mayor, Eric and Knutsen, Dominique},
	month = oct,
	year = {2020},
	pages = {104129},
}

@article{jonnavittula_sari_2024,
	title = {{SARI}: {Shared} {Autonomy} across {Repeated} {Interaction}},
	volume = {13},
	issn = {2573-9522},
	shorttitle = {{SARI}},
	url = {https://dl.acm.org/doi/10.1145/3651994},
	doi = {10.1145/3651994},
	language = {en},
	number = {2},
	urldate = {2025-09-04},
	journal = {ACM Transactions on Human-Robot Interaction},
	author = {Jonnavittula, Ananth and Mehta, Shaunak A. and Losey, Dylan P.},
	month = jun,
	year = {2024},
	pages = {1--36},
}

@article{jung_what_2019,
	title = {What {Do} {Older} {Adults} and {Clinicians} {Think} {About} {Traditional} {Mobility} {Aids} and {Exoskeleton} {Technology}?},
	volume = {8},
	issn = {2573-9522},
	url = {https://dl.acm.org/doi/10.1145/3311789},
	doi = {10.1145/3311789},
	language = {en},
	number = {2},
	urldate = {2025-09-04},
	journal = {ACM Transactions on Human-Robot Interaction},
	author = {Jung, Merel M. and Ludden, Geke D. S.},
	month = jun,
	year = {2019},
	pages = {1--17},
}

@article{lorenzini_ergonomic_2023,
	title = {Ergonomic human-robot collaboration in industry: {A} review},
	volume = {9},
	issn = {2296-9144},
	shorttitle = {Ergonomic human-robot collaboration in industry},
	url = {https://www.frontiersin.org/journals/robotics-and-ai/articles/10.3389/frobt.2022.813907/full},
	doi = {10.3389/frobt.2022.813907},
	language = {English},
	urldate = {2025-09-04},
	journal = {Frontiers in Robotics and AI},
	author = {Lorenzini, Marta and Lagomarsino, Marta and Fortini, Luca and Gholami, Soheil and Ajoudani, Arash},
	month = jan,
	year = {2023},
}

@article{braun_using_2006,
	title = {Using thematic analysis in psychology},
	volume = {3},
	issn = {1478-0887},
	url = {https://doi.org/10.1191/1478088706qp063oa},
	doi = {10.1191/1478088706qp063oa},
	number = {2},
	urldate = {2025-09-03},
	journal = {Qualitative Research in Psychology},
	author = {Braun, Virginia and Clarke, Victoria},
	month = jan,
	year = {2006},
	pages = {77--101},
}

@book{terry_themantic_2017,
	title = {Themantic {Analysis}},
	isbn = {978-1-5264-2286-6},
	language = {en},
	publisher = {SAGE},
	author = {Terry, Gareth and Hayfield, Nikki and Clarke, Victoria and Braun, Virginia},
	month = mar,
	year = {2017},
}

@incollection{braun_doing_2022,
	address = {Cham},
	title = {Doing {Reflexive} {Thematic} {Analysis}},
	isbn = {978-3-031-13942-0},
	url = {https://doi.org/10.1007/978-3-031-13942-0_2},
	doi = {10.1007/978-3-031-13942-0_2},
	language = {en},
	urldate = {2025-09-03},
	booktitle = {Supporting {Research} in {Counselling} and {Psychotherapy} : {Qualitative}, {Quantitative}, and {Mixed} {Methods} {Research}},
	publisher = {Springer International Publishing},
	author = {Braun, Virginia and Clarke, Victoria and Hayfield, Nikki and Davey, Louise and Jenkinson, Elizabeth},
	editor = {Bager-Charleson, Sofie and McBeath, Alistair},
	year = {2022},
	pages = {19--38},
}

@misc{christensen_ordinal_2010,
	title = {ordinal: {Regression} {Models} for {Ordinal} {Data}},
	shorttitle = {ordinal},
	url = {https://CRAN.R-project.org/package=ordinal},
	doi = {10.32614/CRAN.package.ordinal},
	language = {en},
	urldate = {2025-09-03},
	author = {Christensen, Rune Haubo Bojesen},
	month = mar,
	year = {2010},
}

@article{joshi_likert_2015,
	title = {Likert {Scale}: {Explored} and {Explained}},
	volume = {7},
	issn = {22310843},
	shorttitle = {Likert {Scale}},
	url = {https://journalcjast.com/index.php/CJAST/article/view/381},
	doi = {10.9734/BJAST/2015/14975},
	language = {en},
	number = {4},
	urldate = {2025-09-03},
	journal = {British Journal of Applied Science \& Technology},
	author = {Joshi, Ankur and Kale, Saket and Chandel, Satish and Pal, D.},
	month = jan,
	year = {2015},
	pages = {396--403},
}

@article{nomura_measurement_2006,
	title = {Measurement of negative attitudes toward robots},
	volume = {7},
	issn = {1572-0373, 1572-0381},
	url = {https://www.jbe-platform.com/content/journals/10.1075/is.7.3.14nom},
	doi = {10.1075/is.7.3.14nom},
	language = {en},
	number = {3},
	urldate = {2025-09-03},
	journal = {Interaction Studies},
	author = {Nomura, Tatsuya and Suzuki, Tomohiro and Kanda, Takayuki and Kato, Kensuke},
	month = jan,
	year = {2006},
	pages = {437--454},
}

@article{faul_gpower_2007,
	title = {G*{Power} 3: {A} flexible statistical power analysis program for the social, behavioral, and biomedical sciences},
	volume = {39},
	issn = {1554-3528},
	shorttitle = {G*{Power} 3},
	url = {https://doi.org/10.3758/BF03193146},
	doi = {10.3758/BF03193146},
	language = {en},
	number = {2},
	urldate = {2025-09-03},
	journal = {Behavior Research Methods},
	author = {Faul, Franz and Erdfelder, Edgar and Lang, Albert-Georg and Buchner, Axel},
	month = may,
	year = {2007},
	pages = {175--191},
}

@article{pages_tiago_2016,
	title = {{TIAGo}: the modular robot that adapts to different research needs},
	language = {en},
	number = {290},
	journal = {International workshop on robot modularity, IROS},
	author = {Pages, Jordi and Marchionni, Luca and Ferro, Francesco},
	month = oct,
	year = {2016},
	pages = {98},
}

@incollection{creswell_mixed-method_1999,
	address = {San Diego},
	series = {Educational {Psychology}},
	title = {Mixed-{Method} {Research}: {Introduction} and {Application}},
	shorttitle = {Mixed-{Method} {Research}},
	url = {https://www.sciencedirect.com/science/article/pii/B978012174698850045X},
	doi = {10.1016/B978-012174698-8/50045-X},
	urldate = {2025-09-03},
	booktitle = {Handbook of {Educational} {Policy}},
	publisher = {Academic Press},
	author = {Creswell, John W.},
	editor = {Cizek, Gregory J.},
	month = jan,
	year = {1999},
	pages = {455--472},
}

@inproceedings{dai_think_2024,
	title = {Think, {Act}, and {Ask}: {Open}-{World} {Interactive} {Personalized} {Robot} {Navigation}},
	shorttitle = {Think, {Act}, and {Ask}},
	url = {https://ieeexplore.ieee.org/abstract/document/10610178},
	doi = {10.1109/ICRA57147.2024.10610178},
	urldate = {2025-09-02},
	booktitle = {2024 {IEEE} {International} {Conference} on {Robotics} and {Automation} ({ICRA})},
	author = {Dai, Yinpei and Peng, Run and Li, Sikai and Chai, Joyce},
	month = may,
	year = {2024},
	pages = {3296--3303},
}

@inproceedings{whitney_reducing_2017,
	title = {Reducing errors in object-fetching interactions through social feedback},
	url = {https://ieeexplore.ieee.org/abstract/document/7989121},
	doi = {10.1109/ICRA.2017.7989121},
	urldate = {2025-09-02},
	booktitle = {2017 {IEEE} {International} {Conference} on {Robotics} and {Automation} ({ICRA})},
	author = {Whitney, David and Rosen, Eric and MacGlashan, James and Wong, Lawson L. S. and Tellex, Stefanie},
	month = may,
	year = {2017},
	pages = {1006--1013},
}

@article{thierauf_this_2024,
	title = {“{Do} {This} {Instead}”—{Robots} {That} {Adequately} {Respond} to {Corrected} {Instructions}},
	volume = {13},
	url = {https://dl.acm.org/doi/10.1145/3623385},
	doi = {10.1145/3623385},
	language = {en-US},
	number = {3},
	urldate = {2025-09-02},
	journal = {J. Hum.-Robot Interact.},
	author = {Thierauf, Christopher and Thielstrom, Ravenna and Oosterveld, Bradley and Becker, Will and Scheutz, Matthias},
	month = aug,
	year = {2024},
	pages = {40:1--40:23},
}

@inproceedings{bennett_differences_2017,
	title = {Differences in interaction patterns and perception for teleoperated and autonomous humanoid robots},
	issn = {2153-0866},
	url = {https://ieeexplore.ieee.org/abstract/document/8206571},
	doi = {10.1109/IROS.2017.8206571},
	urldate = {2025-09-02},
	booktitle = {2017 {IEEE}/{RSJ} {International} {Conference} on {Intelligent} {Robots} and {Systems} ({IROS})},
	author = {Bennett, Maxwell and Williams, Tom and Thames, Daria and Scheutz, Matthias},
	month = sep,
	year = {2017},
	pages = {6589--6594},
}

@inproceedings{cui_no_2023,
	address = {New York, NY, USA},
	series = {{HRI} '23},
	title = {No, to the {Right}: {Online} {Language} {Corrections} for {Robotic} {Manipulation} via {Shared} {Autonomy}},
	isbn = {978-1-4503-9964-7},
	shorttitle = {No, to the {Right}},
	url = {https://dl.acm.org/doi/10.1145/3568162.3578623},
	doi = {10.1145/3568162.3578623},
	language = {en-US},
	urldate = {2025-09-02},
	booktitle = {Proceedings of the 2023 {ACM}/{IEEE} {International} {Conference} on {Human}-{Robot} {Interaction}},
	publisher = {Association for Computing Machinery},
	author = {Cui, Yuchen and Karamcheti, Siddharth and Palleti, Raj and Shivakumar, Nidhya and Liang, Percy and Sadigh, Dorsa},
	month = mar,
	year = {2023},
	pages = {93--101},
}

@inproceedings{lee_receptionist_2010,
	address = {New York, NY, USA},
	series = {{CSCW} '10},
	title = {Receptionist or information kiosk: how do people talk with a robot?},
	isbn = {978-1-60558-795-0},
	shorttitle = {Receptionist or information kiosk},
	url = {https://dl.acm.org/doi/10.1145/1718918.1718927},
	doi = {10.1145/1718918.1718927},
	urldate = {2025-09-02},
	booktitle = {Proceedings of the 2010 {ACM} conference on {Computer} supported cooperative work},
	publisher = {Association for Computing Machinery},
	author = {Lee, Min Kyung and Kiesler, Sara and Forlizzi, Jodi},
	month = feb,
	year = {2010},
	pages = {31--40},
}

@inproceedings{zhang_can_2025,
	address = {New York, NY, USA},
	series = {{CHI} '25},
	title = {Can you pass that tool?: {Implications} of {Indirect} {Speech} in {Physical} {Human}-{Robot} {Collaboration}},
	isbn = {979-8-4007-1394-1},
	shorttitle = {Can you pass that tool?},
	url = {https://dl.acm.org/doi/10.1145/3706598.3713780},
	doi = {10.1145/3706598.3713780},
	urldate = {2025-09-01},
	booktitle = {Proceedings of the 2025 {CHI} {Conference} on {Human} {Factors} in {Computing} {Systems}},
	publisher = {Association for Computing Machinery},
	author = {Zhang, Yan and Ratnayake, Tharaka Sachintha and Sew, Cherie and Knibbe, Jarrod and Goncalves, Jorge and Johal, Wafa},
	month = apr,
	year = {2025},
	pages = {1--18},
}

@inproceedings{zhang_implicit_2025,
	title = {Implicit {Communication} of {Contextual} {Information} in {Human}-{Robot} {Collaboration}},
	url = {https://ieeexplore.ieee.org/abstract/document/10974036},
	doi = {10.1109/HRI61500.2025.10974036},
	urldate = {2026-01-12},
	booktitle = {2025 20th {ACM}/{IEEE} {International} {Conference} on {Human}-{Robot} {Interaction} ({HRI})},
	author = {Zhang, Yan},
	month = mar,
	year = {2025},
	pages = {1903--1905},
}

@inproceedings{schombs_conversation_2026,
	address = {New York, NY, USA},
	series = {{HAI} '25},
	title = {From {Conversation} to {Orchestration}: {HCI} {Challenges} and {Opportunities} in {Interactive} {Multi}-{Agentic} {Systems}},
	isbn = {979-8-4007-2178-6},
	shorttitle = {From {Conversation} to {Orchestration}},
	url = {https://dl.acm.org/doi/10.1145/3765766.3765795},
	doi = {10.1145/3765766.3765795},
	urldate = {2026-01-11},
	booktitle = {Proceedings of the 13th {International} {Conference} on {Human}-{Agent} {Interaction}},
	publisher = {Association for Computing Machinery},
	author = {Schömbs, Sarah and Zhang, Yan and Goncalves, Jorge and Johal, Wafa},
	month = jan,
	year = {2026},
	pages = {158--168},
}

@article{clark_coordinating_2005,
	title = {Coordinating with each other in a material world},
	volume = {7},
	issn = {1461-4456},
	url = {https://doi.org/10.1177/1461445605054404},
	doi = {10.1177/1461445605054404},
	language = {EN},
	number = {4-5},
	urldate = {2026-01-28},
	journal = {Discourse Studies},
	publisher = {SAGE Publications},
	author = {Clark, Herbert H.},
	month = oct,
	year = {2005},
	pages = {507--525},
}

@article{ananth_regression_1997,
	title = {Regression models for ordinal responses: a review of methods and applications.},
	volume = {26},
	issn = {0300-5771},
	shorttitle = {Regression models for ordinal responses},
	url = {https://doi.org/10.1093/ije/26.6.1323},
	doi = {10.1093/ije/26.6.1323},
	number = {6},
	urldate = {2026-01-18},
	journal = {International Journal of Epidemiology},
	author = {Ananth, C V and Kleinbaum, D G},
	month = dec,
	year = {1997},
	pages = {1323--1333},
}

@inproceedings{kiesler_fostering_2005,
	title = {Fostering common ground in human-robot interaction},
	issn = {1944-9437},
	url = {https://ieeexplore.ieee.org/abstract/document/1513866},
	doi = {10.1109/ROMAN.2005.1513866},
	urldate = {2025-08-31},
	booktitle = {{ROMAN} 2005. {IEEE} {International} {Workshop} on {Robot} and {Human} {Interactive} {Communication}, 2005.},
	author = {Kiesler, S.},
	month = aug,
	year = {2005},
	pages = {729--734},
}

@inproceedings{williams_thank_2018,
	address = {New York, NY, USA},
	series = {{HRI} '18},
	title = {"{Thank} {You} for {Sharing} that {Interesting} {Fact}!": {Effects} of {Capability} and {Context} on {Indirect} {Speech} {Act} {Use} in {Task}-{Based} {Human}-{Robot} {Dialogue}},
	isbn = {978-1-4503-4953-6},
	shorttitle = {"{Thank} {You} for {Sharing} that {Interesting} {Fact}!"},
	url = {https://dl.acm.org/doi/10.1145/3171221.3171246},
	doi = {10.1145/3171221.3171246},
	urldate = {2026-01-11},
	booktitle = {Proceedings of the 2018 {ACM}/{IEEE} {International} {Conference} on {Human}-{Robot} {Interaction}},
	publisher = {Association for Computing Machinery},
	author = {Williams, Tom and Thames, Daria and Novakoff, Julia and Scheutz, Matthias},
	month = feb,
	year = {2018},
	pages = {298--306},
}

@article{reeves_opening_2025,
	title = {Opening {Up} {Human}-{Robot} {Collaboration}},
	volume = {9},
	url = {https://doi.org/10.1145/3757640},
	doi = {10.1145/3757640},
	language = {en},
	number = {7},
	journal = {Proceedings of the ACM on Human-Computer Interaction},
	author = {Reeves, Stuart and Pelikan, Hannah R M and Cantarutti, Marina N},
	year = {2025},
}

@article{liu_review_2019,
	title = {A review of methodologies for natural-language-facilitated human–robot cooperation},
	volume = {16},
	issn = {1729-8806},
	url = {https://doi.org/10.1177/1729881419851402},
	doi = {10.1177/1729881419851402},
	language = {EN},
	number = {3},
	urldate = {2025-09-02},
	journal = {International Journal of Advanced Robotic Systems},
	author = {Liu, Rui and Zhang, Xiaoli},
	month = may,
	year = {2019},
	pages = {1729881419851402},
}

@misc{robotics_tiago_nodate,
	title = {{TIAGo} {Head} {\textbar} {The} {Ultimate} {Platform} for {HRI} and {Embodied} {AI}},
	url = {https://pal-robotics.com/robot/tiago-head/},
	language = {en-US},
	urldate = {2025-09-10},
	journal = {PAL Robotics},
	author = {Robotics, PAL},
}

@inproceedings{lyu_signaling_2025,
	address = {New York, NY, USA},
	series = {{CHI} '25},
	title = {Signaling {Human} {Intentions} to {Service} {Robots}: {Understanding} the {Use} of {Social} {Cues} during {In}-{Person} {Conversations}},
	isbn = {979-8-4007-1394-1},
	shorttitle = {Signaling {Human} {Intentions} to {Service} {Robots}},
	url = {https://dl.acm.org/doi/10.1145/3706598.3714235},
	doi = {10.1145/3706598.3714235},
	urldate = {2025-09-07},
	booktitle = {Proceedings of the 2025 {CHI} {Conference} on {Human} {Factors} in {Computing} {Systems}},
	publisher = {Association for Computing Machinery},
	author = {Lyu, Hanfang and Wang, Xiaoyu and Zhang, Nandi and Ma, Shuai and Zhu, Qian and Luo, Yuhan and Tsung, Fugee and Ma, Xiaojuan},
	month = apr,
	year = {2025},
	pages = {1--21},
}

@inproceedings{hough_its_2017,
	address = {New York, NY, USA},
	series = {{HRI} '17},
	title = {It's {Not} {What} {You} {Do}, {It}'s {How} {You} {Do} {It}: {Grounding} {Uncertainty} for a {Simple} {Robot}},
	isbn = {978-1-4503-4336-7},
	shorttitle = {It's {Not} {What} {You} {Do}, {It}'s {How} {You} {Do} {It}},
	url = {https://dl.acm.org/doi/10.1145/2909824.3020214},
	doi = {10.1145/2909824.3020214},
	language = {en-US},
	urldate = {2025-09-07},
	booktitle = {Proceedings of the 2017 {ACM}/{IEEE} {International} {Conference} on {Human}-{Robot} {Interaction}},
	publisher = {Association for Computing Machinery},
	author = {Hough, Julian and Schlangen, David},
	month = mar,
	year = {2017},
	pages = {274--282},
}

\appendix
\onecolumn
\section*{Appendix}
\renewcommand{\thetable}{A\arabic{table}}
\setcounter{table}{0} 

\section{The Full Results of CLMM}

\begin{table}[H]
\centering
\small
\caption{Model summaries for CLMM analysis}
\label{tab:clmm_all}
\begin{tabular}{|lrrrrr|}
\hline
\multicolumn{6}{|c|}{\cellcolor[HTML]{FDEAE3}\textbf{Grounding with Collaborator}}\\
\hline
\textbf{Predictor} & \textbf{Estimates ($\beta$)} & \textbf{Effect Sizes (OR)} & \textbf{Std. Error} & \textbf{95\% CI} & \textbf{p-value}\\
\hline
Explicit                    &  2.677  & 14.54 & 1.610 & -0.476 -- 5.830    & 0.096. \\
Incorrect                   & -10.737 & 46027.77 & 1.950 & -14.558 -- -6.916   & \textless{}0.001* \\
High sev.                   & -0.434  & 1.54 & 0.838 & -2.077 -- 1.208   & 0.604 \\
Prev.collab.               &  0.302  & 1.35 & 0.378 & -0.439 -- 1.043   & 0.425 \\
Explicit $\times$ High      & -1.334  & 3.80 & 0.925 & -3.147 -- 0.478   & 0.149 \\
Incorrect $\times$ High     &  0.967  & 2.63 & 1.006 & -1.005 -- 2.939    & 0.337 \\
Explicit $\times$ Incorrect & -2.000  & 7.39 & 1.708 & -5.348 -- 1.348   & 0.242 \\
\hline
\textbf{Random Effects} & \textbf{Variance} & \textbf{Std. Dev} & \textbf{Correlation} & &\\
\hline
pid & 4.899 & 2.213 &  & &\\
timing 2 $\mid$ pid & 1.173 & 1.083 & -0.834 && \\
timing 3 $\mid$ pid & 1.670 & 1.292 & -0.579 & 0.034 & \\
\hline
\textbf{Model Fit} & \textbf{AIC} & \textbf{Log Lik} & & & \\
\hline
 & 255.060 & -109.530 &  && \\
\hline
\multicolumn{6}{|l|}{$ Variable\sim Explicitness\times Correctness + Correctness\times Severity +Explicitness\times Severity$} \\ 
&\multicolumn{5}{l|}{$+ Prev.collab. + (1+timing\mid pid) + (1\mid timing)$}\\
\hline

\multicolumn{6}{|c|}{\cellcolor[HTML]{FDEAE3}\textbf{Human Communication Effort}}\\
\hline
\textbf{Predictor} & \textbf{Estimates ($\beta$)} & \textbf{Effect Sizes (OR)} & \textbf{Std. Error} & \textbf{95\% CI} & \textbf{p-value}\\
\hline
Explicit                    &  1.200 & 3.32 & 0.002 & 1.195 -- 1.204     & \textless{}0.001* \\
Incorrect                   &  3.910 & 49.90 & 1.623 & 0.728 -- 7.092     & 0.016* \\
High sev.                   &  0.269 & 1.31 & 0.002 & 0.264 -- 0.273     & \textless{}0.001* \\
Prev.collab.               & -0.112 & 1.12 & 0.002 & -0.116 -- -0.107   & \textless{}0.001* \\
Explicit $\times$ High      & -0.249 & 1.28 & 0.002 & -0.253 -- -0.244   & \textless{}0.001* \\
Incorrect $\times$ High     &  0.376 & 1.46 & 0.805 & -1.202 -- 1.954    & 0.640 \\
Explicit $\times$ Incorrect & -3.632 & 37.79  & 1.940 & -7.434 -- 0.170     & 0.061. \\
\hline
\textbf{Random Effects} & \textbf{Variance} & \textbf{Std. Dev} &  && \\
\hline
pid & 9.136 & 3.023 &  & &\\
\hline
\textbf{Model Fit} & \textbf{AIC} & \textbf{Log Lik} &  & &\\
\hline
 & 169.530 & -72.770 &  & &\\
\hline
\multicolumn{6}{|l|}{$Variable\sim  Explicitness\times Correctness   + Correctness\times Severity +Explicitness\times Severity$} \\ 
&\multicolumn{5}{l|}{$+ Prev.collab. + (1\mid pid)$}\\
\hline

\multicolumn{6}{|c|}{\cellcolor[HTML]{FDEAE3}\textbf{Collaborator Communication Effort}}\\
\hline
\textbf{Predictor} & \textbf{Estimates ($\beta$)} & \textbf{Effect Sizes (OR)}& \textbf{Std. Error} & \textbf{95\% CI} & \textbf{p-value}\\
\hline
Explicit                    & -0.108 & 1.11 & 1.440 & -2.930 -- 2.715     & 0.940 \\
Incorrect                   & -2.322 & 10.20 & 1.481 & -5.224 -- 0.580       & 0.117 \\
High sev.                   &  1.776 & 5.91 & 0.932 & -0.050 -- 3.603    & 0.057. \\
Prev.collab.               &  0.123 & 1.13  & 0.418 & -0.695 -- 0.942      & 0.768 \\
Explicit $\times$ High      & -1.735 & 5.67 & 1.046 & -3.785 -- 0.316   & 0.097. \\
Incorrect $\times$ High     & -1.702 & 5.49   & 1.047 & -3.755 -- 0.350       & 0.104 \\
Explicit $\times$ Incorrect &  4.728 & 113.07  & 2.050 & 0.710 -- 8.745        & 0.021* \\
\hline
\textbf{Random Effects} & \textbf{Variance} & \textbf{Std. Dev} &  && \\
\hline
pid & 5.109 & 2.260 & & & \\
\hline
\textbf{Model Fit} & \textbf{AIC} & \textbf{Log Lik} &&  & \\
\hline
 & 185.430 & -80.720 & & & \\
 \hline
\multicolumn{6}{|l|}{$Variable \sim Explicitness\times Correctness + Correctness\times Severity + Explicitness\times Severity $} \\ 
&\multicolumn{5}{l|}{$+ Prev.collab. + (1\mid pid)$}\\
\hline
\end{tabular}
\end{table}

\newpage

\section{Selected Questions from the Negative Attitude Towards Robots Scale} \label{apd:NARS}
``Please read through each of the following sentences and then indicate how frequently these statements apply to you, from (1) strongly disagree to (5) strongly agree.''
\begin{itemize}
    \item I would feel very nervous just standing in front of a robot.
    \item I would feel very nervous talking with robots.
    \item I would feel uneasy if I were given a job where I had to use robots.
\end{itemize}

\section{Questionnaires Questions} \label{apd:questionnaire}
\begin{itemize}
    \item Perceived Grounding with the Collaborator: To what extent do you feel you and Tiago had a mutual understanding at this moment? (1: Not at all; 5: Completely)
    \item Human Communication Effort: Please rate how much effort you put into communication during the task. (1: None at all; 5: A great deal)
    \item Collaborator Communication Effort: Please rate how much effort Tiago put into communication during the task. (1: None at all; 5: A great deal)
\end{itemize}

\section{Example of the Interaction between the Collaborator and the Facilitator} \label{apd:example}

\begin{table}[h]
\centering
\caption{An example of Facilitator's indirect requests and Collaborator's responses.}
\resizebox{1\textwidth}{!}{
\begin{tabular}{|l|l|llll|}
    \hline
    \multicolumn{6}{|l|}{\begin{tabular}[c]{@{}l@{}}\textbf{Context}: In the medical task, the Collaborator is holding the test tube with liquid inside, the participant is pouring powder into the test tube.\end{tabular}}\\
    \hline
    \hline
    \textbf{Dialogue turn} & \textbf{Facilitator} & \textbf{IC Collaborator} & \textbf{II Collaborator} & \textbf{EC Collaborator} & \textbf{EI Collaborator} \\
    \hline
    Turn 1 & ``That won't mix itself.'' & \begin{tabular}[c]{@{}l@{}}``Got it.\\ I'll shake the test tube.'' \\(Shaking the test tube) \end{tabular} &  \begin{tabular}[c]{@{}l@{}} ``Sounds like we did it wrong. \\Let's start over.''\\ (Placing the test tube back) \end{tabular} & \begin{tabular}[c]{@{}l@{}}``Do you want me to \\shake the test tube?''\end{tabular} & \begin{tabular}[c]{@{}l@{}}``Did we do it wrong? \\Should we start over?''\end{tabular}\\
    \hline
    Turn 2A & ``Yes.'' & -- & -- & \begin{tabular}[c]{@{}l@{}}``Got it.''\\ (Shaking the test tube) \end{tabular} & --\\
    \hline
    Turn 2B & \begin{tabular}[c]{@{}l@{}}``No. I mean, \\please shake the test tube.''\end{tabular} & -- & \begin{tabular}[c]{@{}l@{}}``Got it.''\\ (Pick \& shake the test tube) \end{tabular} & -- & \begin{tabular}[c]{@{}l@{}}``Got it.''\\ (Shaking the test tube) \end{tabular}\\
    \hline
\end{tabular}
}
\end{table}

\end{document}